\documentclass[10pt,twocolumn,letterpaper]{article}
\pdfoutput=1
\usepackage{iccv}
\usepackage{times}
\usepackage{epsfig}
\usepackage{graphicx}
\usepackage{tabularx}
\usepackage[caption=false]{subfig}
\usepackage{amsmath}
\usepackage{overpic}
\usepackage{amssymb}

\newcommand{\boldparagraph}[1]{\vspace{0.0em}\noindent{\bf #1} }

\newcommand{\sV}{\mathcal{V}}

\usepackage{multirow}
\usepackage{makecell}

\graphicspath{{figures/} {exp_imgs/}}
\usepackage[colorlinks,bookmarks=false]{hyperref}

\iccvfinalcopy 

\def\iccvPaperID{5970} 

\ificcvfinal\fi

\begin{document}

\title{Vis2Mesh: Efficient Mesh Reconstruction from Unstructured Point Clouds of Large Scenes with Learned Virtual View Visibility}

\author{Shuang Song\\
The Ohio State University
\and
Zhaopeng Cui\thanks{Contribution initiated and performed at ETH Zürich.}\\
Zhejiang University
\and
Rongjun Qin\thanks{Corresponding author. Email: \small \href{mailto:qin.324@osu.edu}{qin.324@osu.edu}}\\
The Ohio State University
}

\maketitle
\ificcvfinal\thispagestyle{empty}\fi
\begin{abstract}
 We present a novel framework for mesh reconstruction from unstructured point clouds by taking advantage of the learned visibility of the 3D points in the virtual views and traditional graph-cut based mesh generation. Specifically, we first propose a three-step network that explicitly employs depth completion for visibility prediction. 
 Then the visibility information of multiple views is aggregated to generate a 3D mesh model by solving an optimization problem considering visibility in which a novel adaptive visibility weighting in surface determination is also introduced to suppress line of sight with a large incident angle. Compared to other learning-based approaches, our pipeline only exercises the learning on a 2D binary classification task, \ie, points visible or not in a view, which is much more generalizable and practically more efficient and capable to deal with a large number of points. Experiments demonstrate that our method with favorable transferability and robustness, and achieve competing performances \wrt state-of-the-art learning-based approaches on small complex objects and outperforms on large indoor and outdoor scenes. Code is available at \href{https://github.com/GDAOSU/vis2mesh}{https://github.com/GDAOSU/vis2mesh}.
\end{abstract}
\vspace{-2em}
\section{Introduction}
\begin{figure}[t!]
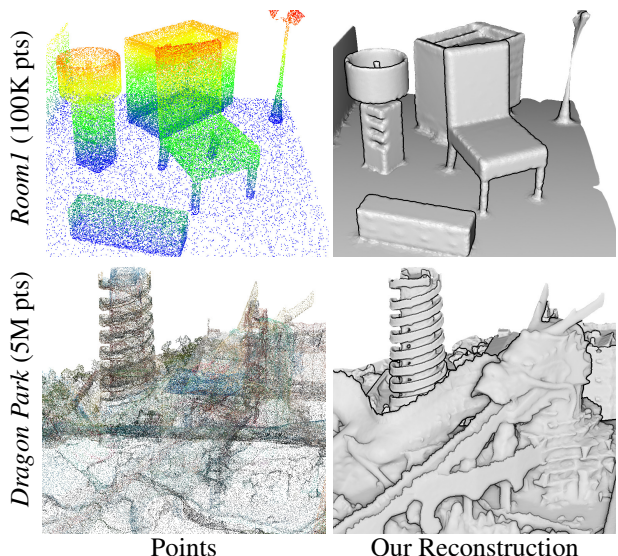

    \centering
    \def\subimgwidth{0.45\linewidth}
    \edef\roomOneViewTwoopts{trim=300 20 230 20,clip,width = \subimgwidth}
    \edef\dragonTsViewopts{trim=300 30 300 30,clip,width = \subimgwidth}
    
    \setlength{\tabcolsep}{1pt}
    \def\arraystretch{1}
    
    \begin{tabular}{ccc}
        \rotatebox[origin=c]{90}{\textit{Room1} (100K pts)} &
        \raisebox{-0.5\height}{
         \expandafter\includegraphics\expandafter[\roomOneViewTwoopts]{indoor_room_1_raw/pointcloud_ramp_view_2.jpg}} & \raisebox{-0.5\height}{\expandafter\includegraphics\expandafter[\roomOneViewTwoopts]{indoor_room_1_raw/ours_angle_view_2.jpg}} \\
         \rotatebox[origin=c]{90}{\textit{Dragon Park} (5M pts)} &
         \raisebox{-0.5\height}{\expandafter\includegraphics\expandafter[\dragonTsViewopts]{dragonpark_5M/point_sparse_view3.jpg}} & 
         \raisebox{-0.5\height}{\expandafter\includegraphics\expandafter[\dragonTsViewopts]{dragonpark_5M/ours_angle_view3.jpg}} \\
         & Points & Our Reconstruction
    \end{tabular}

    \caption{Examples of reconstructed surfaces with our approach on both indoor and large outdoor scenes. The number of points ranges from thousands to millions.}
    \label{fig:teaser}
\end{figure}

3D surface reconstruction is essential to drive many computer vision and VR/AR applications, such as vision-based localization, view rendering, animation, and autonomous navigation. With the development of 3D sensors, \eg, LiDAR or depth sensors, we can directly capture accurate 3D point clouds.
However, determining surfaces from unstructured point clouds remains a challenging problem, especially for point clouds of objects of complex shapes, varying point density, completeness, and volumes. A favorable mesh reconstruction method should be 1) capable of recovering geometric details captured by the point clouds; 2) robust and generalizable to different scene contexts; 3) scalable to large scenes and tractable in terms of memory and computation.


Typical reconstruction methods either explicitly explore the local surface recovery through connecting neighboring points (\eg Delaunay Triangulation~\cite{delaunayChapter}), or sort solutions globally through implicit surface determination (\eg Screened Poisson Surface Reconstruction~\cite{kazhdan2006poisson,kazhdan2013screened} ). However, both of them favor high-density point clouds. With the development of 3D deep learning, many learning-based methods have demonstrated their strong capability when reconstructing mesh surfaces from moderately dense or even comparatively sparse point clouds of complex objects. However, these end-to-end deep architectures, on one hand, encode the contextual/scene information which often runs into generalization issues; on the other hand, the heavy networks challenge the memory and computations when scaling up to large scenes. 

The point visibility in views has been shown as a good source of information, which can greatly benefit surface reconstruction~\cite{song2020optimizing}. For example, a 3D point visible on a physical image view alludes to the fact the line of sight from the 3D point to the view perspective center, travels through free space (no objects or parts occludes), which as a result, can serve as a strong constraint to guide the mesh reconstruction. Intuitively, the more visible a point is in more views, the more information about the free space between points and views that one can explore~\cite{weaklysupportCVPR11}. This principle has been practiced in multi-view stereo (MVS) and range images and shows promising results~\cite{labatut2007efficient,labatut2009robust,highaccrecon2012}, in which the visibility of each 3D point to each physical is recorded through dense matching (in MVS) or decoded directly from the sensors. However, in both cases, the visibility is limited by the number of physical views and accounts for only limited knowledge of the free space, thus leading to the lack of reconstruction details.
%

To address it, we propose a novel solution by generating a large number of virtual views around the point clouds and utilize the learned visibility through a graph-cut based surface reconstruction approach. This will literally give us unlimited visibility information for high-quality mesh generation through traditional approaches based on visibility. However, it is non-trivial to design such a system. First, visibility prediction in virtual views is not easy as the projected pixels from the point clouds to different views can have varying sparsity. Second, unlike the MVS views which normally assume a good incidence angle to the generated point clouds, the generated virtual views may present a large variation in terms of their incident angle and distance to the surface. 
To solve these problems, we design a network for visibility prediction, which utilizes partial convolution to model the sparsity of the input in the neural network. To further improve the accuracy, we propose a cascade network that explicitly employs depth completion as an intermediate task. Secondly, to suppress the side effect of unfavored virtual rays (large-incidence angle to the surface), we propose a novel adaptive visibility weighting term that adaptively weights the virtual rays. Compared to the end-to-end learning-based methods, our method has better generalization across different scene contexts because a very simple and learning-friendly task, 
\ie, visibility prediction, is involved in our pipeline. Moreover, our method also maintains the capability to process an extremely large volume of outdoor point clouds with high quality.



Our \textbf{contributions} can be summarized as follows: Firstly, we present a surface reconstruction framework from unstructured point clouds, that exploits the power of both traditional and learning-based methods; Secondly, we propose a three-step network that explicitly employs the depth completion for the visibility prediction of 3D points; Thirdly, we propose a novel adaptive visibility weighting term into the traditional graph-cut based meshing pipeline, to increase the geometric details; Lastly, we demonstrate through our experiments that our proposed method yields better generalization capability, and can be scaled up to process point clouds of a large scene. It is also robust against different types of noises and incomplete data and has better performance than the state-of-the-art methods.

\begin{figure*}
    \centering
    \includegraphics[width=\linewidth]{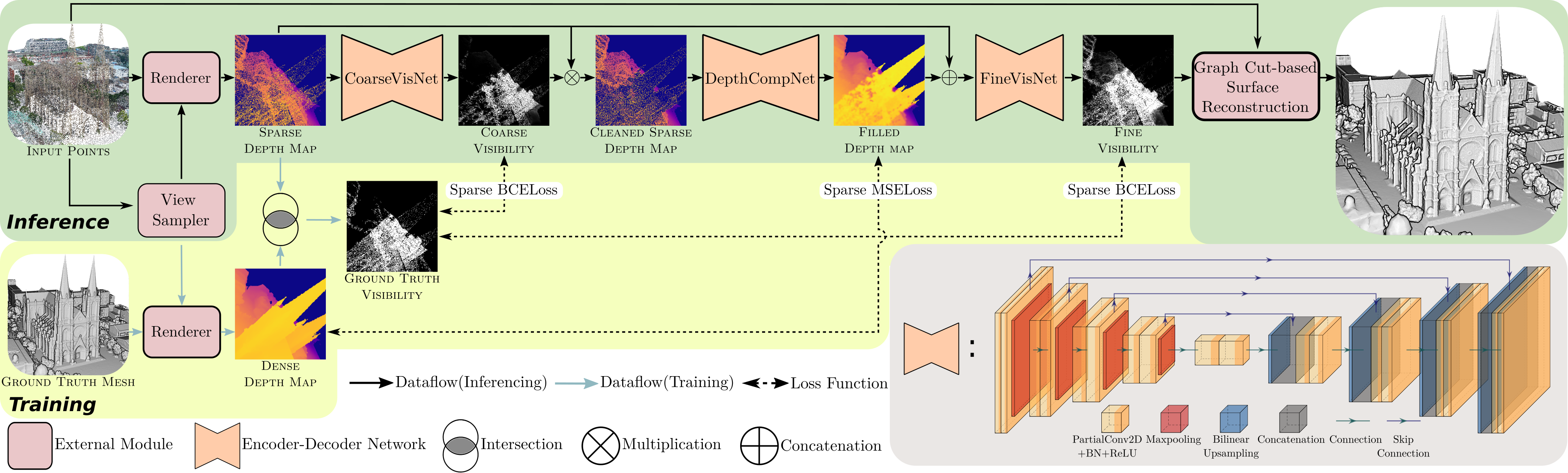}
    \caption{The proposed Vis2Mesh framework. Our framework reconstructs surfaces in \textit{four} steps shown in inference region from left to right follow the flow of arrows: 1) Virtual view sampling 2) Rendering 3) Visibility determination and 4) Delaunay and graph-cut based surface reconstruction. Visibility determination composed by three networks: \textbf{CoarseVisNet}, predicting visibility from sparse depth input, \textbf{DepthCompNet} completing dense depth map based on coarse predicted visible points and \textbf{FineVisNet} refining visibility prediction with sparse depth and completed dense depth map, detailed in Section~\ref{sec:method:networks}.}
    \label{fig:architecture}
\end{figure*}

\section{Related Work}


\boldparagraph{Classic surface reconstruction} from unstructured point clouds exploits both local and global surface smoothness, global regularity, visibility, \etc~\cite{survey2017pointsurfacerecon}. The \textit{local surface smoothness} based methods~\cite{edelsbrunner1994three, hoppe1992surface, alexa2003computing} seek for smooth surface only in close proximity to the data.
The \textit{global surface smoothness} based methods~\cite{kazhdan2005reconstruction, carr2001reconstruction, kazhdan2006poisson, kazhdan2013screened, kolluri2004spectral} tend to find a field function (signed distance function, indicator function) approximating the point cloud. The most common global smoothness algorithm is Screened Poisson Surface Reconstruction (SPSR)~\cite{kazhdan2006poisson, kazhdan2013screened} which can generate smooth, void-free surfaces with high-quality oriented normals, while it can be oversensitive to incorrect normals, as well as non-watertight shapes.
The \textit{global regularity} based methods~\cite{oesau2016planar, zheng2010non,li20112d} deal with the man-made objects or architectural shapes that feature symmetry, orthogonality, repetition, and parallelism.
The \textit{visibility} based methods assume the information of the sensors from which the point clouds are collected, and this yields additional cues as rays and visibility of each point at sensor locations in which they are observable. Varying forms of visibility (includes lines of sight~\cite{tsdf96, labatut2007efficient}, exterior visibility~\cite{shalom2010cone, katz2007direct}, parity~\cite{nagai2015tomographic}) have been explored. The most widely used methods utilize lines of sight~\cite{labatut2007efficient, labatut2009robust, weaklysupportCVPR11, weaklysupport2014, zhou2019detail} that determine the surface between interior and exterior tetrahedron of the Delaunay triangulation of a point set by solving a \textit{s-t} graph which is weighted by lines of sight. Nonetheless, the quality of results relies on the number and quality of rays and it may fall short in concave areas on the reconstructed surface if insufficient rays are identified~\cite{zhou2019detail}.

\boldparagraph{Learning-based surface reconstruction} takes a data-driven approach that takes advantage of the available examples and the high-capability network structures~\cite{qi2017pointnet, qi2017pointnet++, 3dunet, dgcnn} to build surfaces for challenging and complex point clouds that classic methods have poor performance on. Several methods~\cite{chen2019learning, mescheder2019occupancy, park2019deepsdf} encode the entire object to an embedding vector and decode it with a neural network, as a general problem in such deep networks, the representation often cannot be generalized to objects of unseen categories or scenes. CONet~\cite{Peng2020ECCV} proposed alternative volumetric representations and enabled convolution in the implicit field making the representations invariant to translation, and the sliding-window nature of this approach makes CONet capable of scaling up to large scenes, while the method being local does not capture global shape information and the volumetric representations are still memory demanding for 3D data generation.  Points2Surf~\cite{ErlerEtAl:Points2Surf:ECCV:2020} proposed a patch-based learning framework combining outputs of local-scale and global-scale neural networks to construct a signed distance function (SDF) and shows a certain level of generalization capability. Yet the query on the per-voxel level and the inference over the voxels field are computationally intensive (4x slower than SPSR), which limits it to only process small scenes.
The MIER method~\cite{liu2020meshing} uses the explicit representation of input points and cast the connectivity to a surrogate comparing geodesic/euclidean distances. A neural network-based classifier is trained to determine surface triangles, which has shown to be capable of preserving fine-grained details and works well on ambiguous structures. However, the per-triangle classification overlooks the connectivity of adjacent triangles, causing holes on the reconstructed surfaces.
Point2Mesh~\cite{Hanocka2020p2m} proposed a self-prior to detect self-similarity, which contributes to reconstructing surface as well as removing noises and completing missing parts. Nonetheless, Point2Mesh only works with watertight and simple models, requires an initial mesh model, and comparably needs a long running time to converge. As compared to the classic methods, the learning-based methods, in general, demand a high volume of memory and can be scene-specific.

\section{Preliminary\label{sec:prelimiary}}
This section introduces the formula of Delaunay triangulation and graph-cut optimization based surface reconstruction~\cite{labatut2009robust}.
The surface reconstruction problem is considered as the outer space/inner space labeling problem of the connected graph built based on Delaunay triangulation, to be solved by minimum s-t cut algorithm~\cite{boykov2001fast}. The reconstructed surface, denoted by $S$ in the following, is a union of oriented Delaunay triangles which are guaranteed to be watertight and intersection free as it bounds a volume consist of tetrahedra. The energy function of the reconstruction problem is formulated by the basic method~\cite{labatut2009robust} as Equation~\ref{eq:globalenergy}.
\begin{equation}
\small
\vspace{-0.5em}
\begin{split}
    E(S) &= E_{vis}(S,\sV) + \lambda_{ql}E_{ql}(S) \\
    &=\sum_{v \in \sV}[E_{s}(v) + E_{t}(v) + E_{ij}(v)] + \lambda_{ql}E_{ql}(S),
\end{split}
    \label{eq:globalenergy}
\end{equation}
where $E_{vis}(S,\sV)$ is a sum of penalties for conflicts and wrong orientations of the surface $S$ concerning the constraints imposed by all lines of sight $\sV$, $E_{ql}(S)$ is an additional smoothness term to suppress skinny triangles~\cite{labatut2009robust}. In graph-cut, $E_{s}$ and $E_{t}$ are data terms for outer and inner space respectively, and $E_{ij}$ is a smoothness term. $E_{ql}(S)$ is another smoothness term independent of visibility.
\begin{subequations}
\small
\begin{align}
   & E_{s}(v) = \infty \cdot \delta[T_1 \in t] \label{eq:origin_E_s},\\
   & E_{t}(v) = \alpha_{vis} \cdot \delta[T_{M+1} \in s] \label{eq:origin_E_t},\\
   & E_{ij}(v) = \sum_{i=1}^{M-1} \alpha_{vis} \cdot \delta[T_i \in s \wedge T_{i+1} \in t], \label{eq:origin_E_ij}
\end{align}
\label{eq:origin_terms}
\end{subequations}
where $T_i$ is the $i$-th tetrahedron intersects with the line of sight $v$, $M$ is the size of the intersection set. So that $T_1$ and $T_{M+1}$ denote the first and the one right behind the last tetrahedra along $v$. The conditional operator $\delta[\cdot]$ is 1 if the condition is true, otherwise is 0. $\alpha_{vis}$ is the confidence of the visibility which is proportional to the distance to the endpoint, which is called soft visibility constraint. Please refer to our supp. material for more details about graph weighting.
\section{Method}
In this section, we introduce our framework of surface reconstruction from unstructured point clouds  (Figure~\ref{fig:architecture}). 
Firstly, we sample feasible 6 DoF (Degree of freedom) poses according to input point clouds. Then, a 3D renderer will take poses as viewing parameters and project points onto that virtual image plane based on its projection matrix. Next, a three-step network that explicitly employs depth completion to assist visibility estimation, and predicts the visible and occluded points to build visibility information. Finally, the graph-cut based solver processes the input points associated with predicted visibility to reconstruct the surface. Our overall framework is shown in Figure~\ref{fig:architecture}, and the individual components will be discussed in the following subsections.

\subsection{Virtual View Sampling\label{sec:viewgeneratos}}
\begin{figure}
    \centering
    \def\subimgwidth{0.24\linewidth}
    
    \setlength\tabcolsep{1pt}
    \def\arraystretch{0}
    \begin{tabular}{cccc}
    \subfloat[Spherical]{%
        \includegraphics[trim=340 50 340 50,clip, width=\subimgwidth]{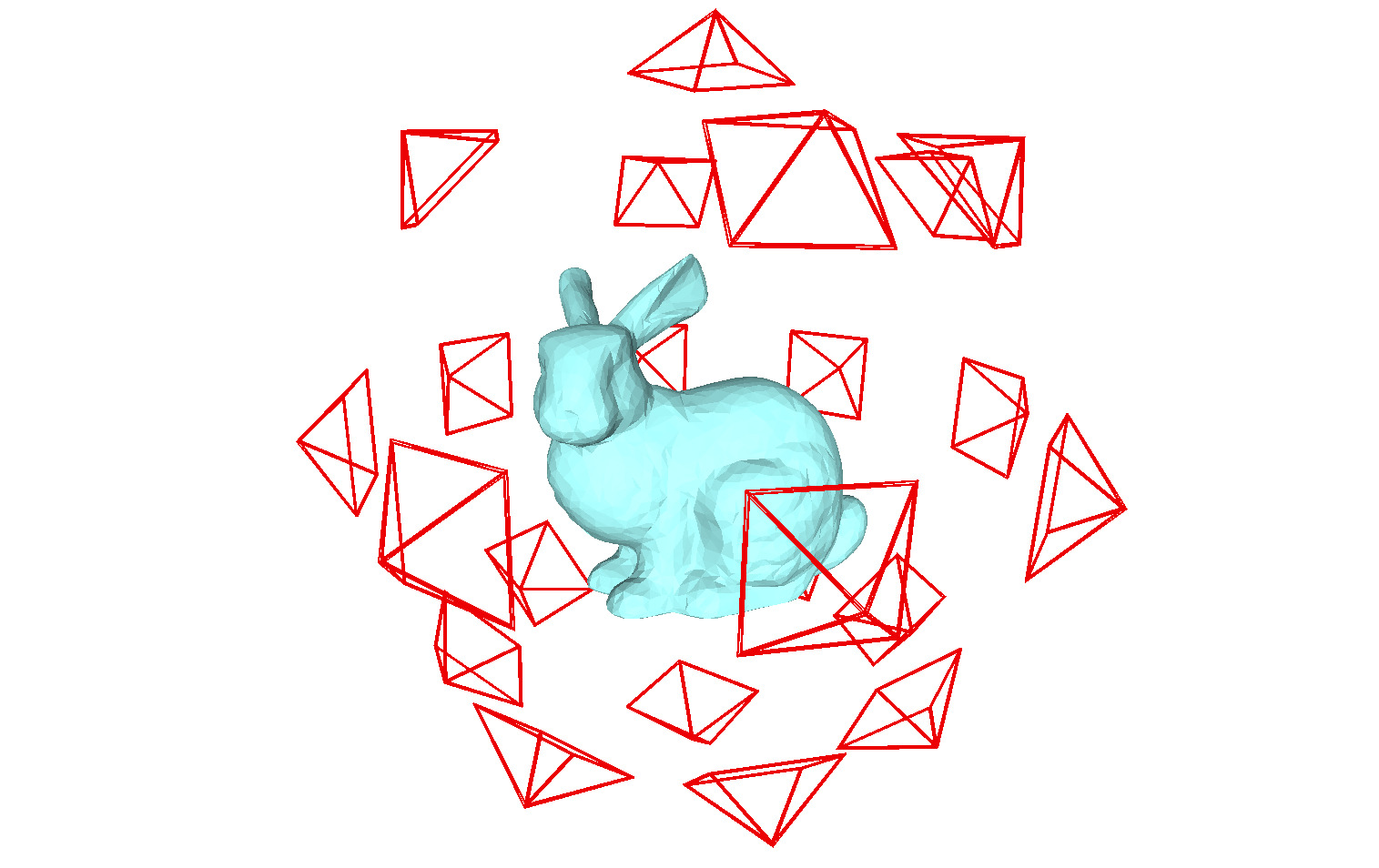}}  & 
    \subfloat[Grid+Nadir]{%
        \includegraphics[trim=340 50 340 50,clip,width=\subimgwidth]{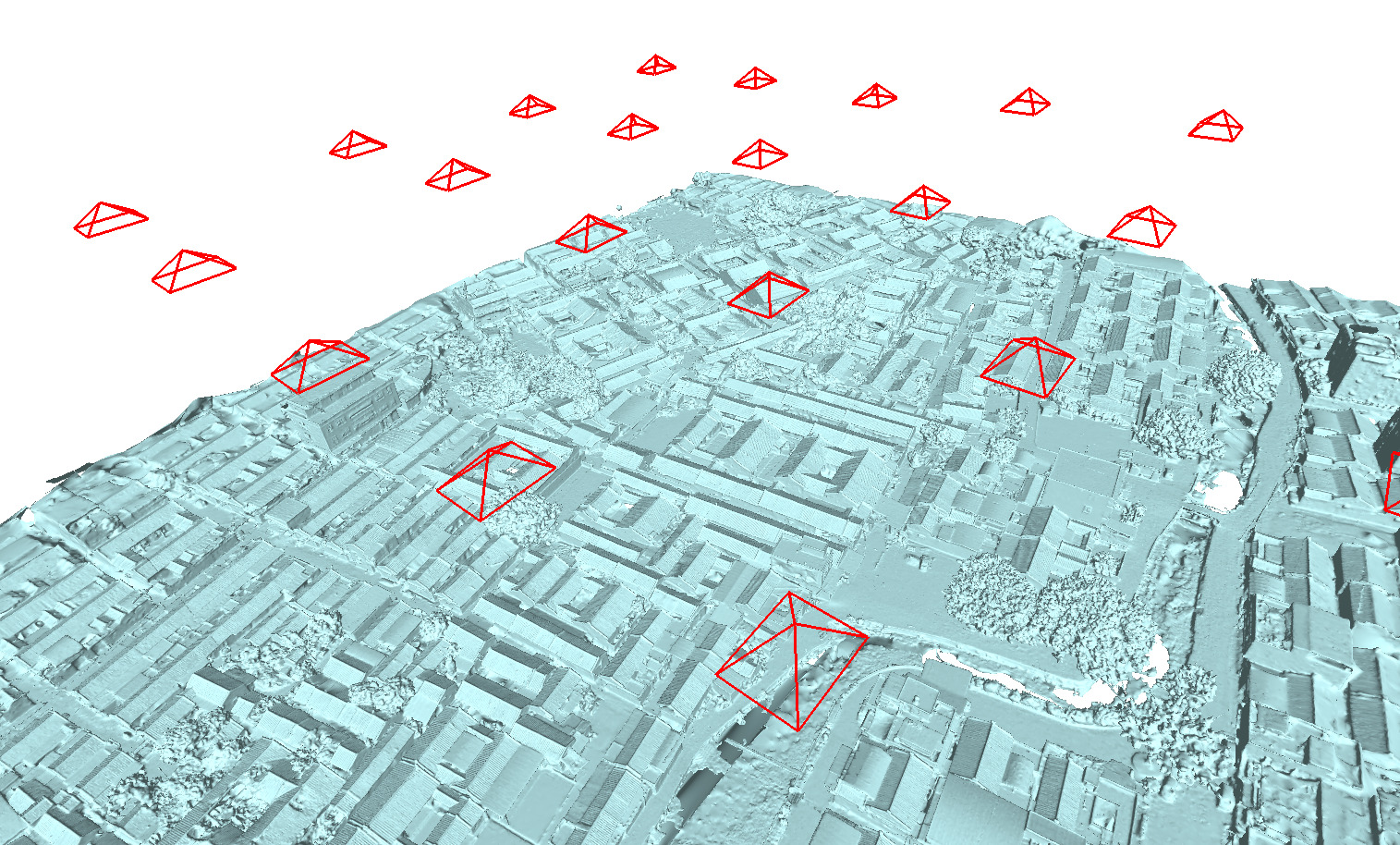}} &
    \subfloat[Grid+Oblique]{%
        \includegraphics[trim=340 50 340 50,clip,width=\subimgwidth]{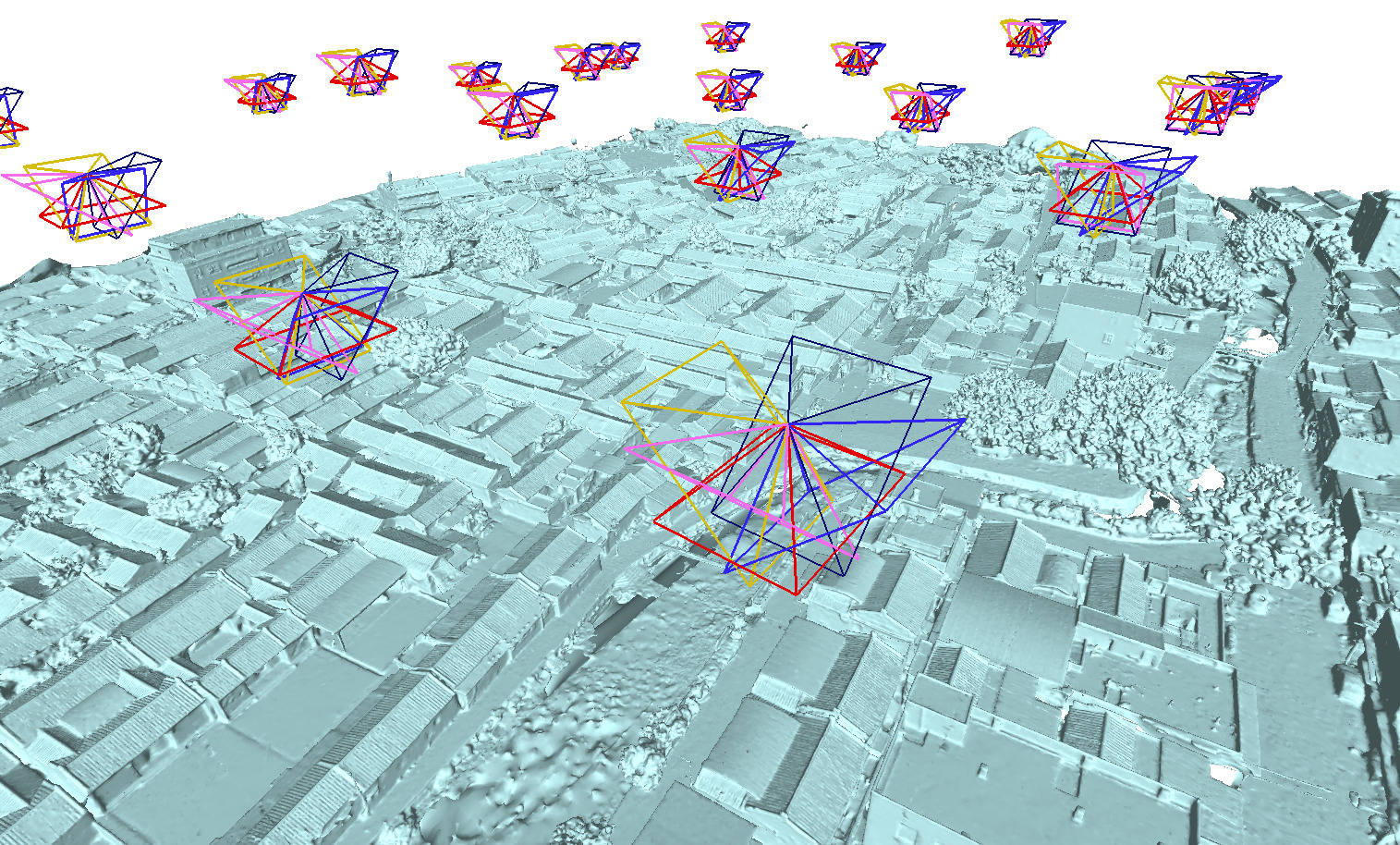}} &
    \subfloat[Customized]{%
        \includegraphics[trim=340 50 340 50,clip,width=\subimgwidth]{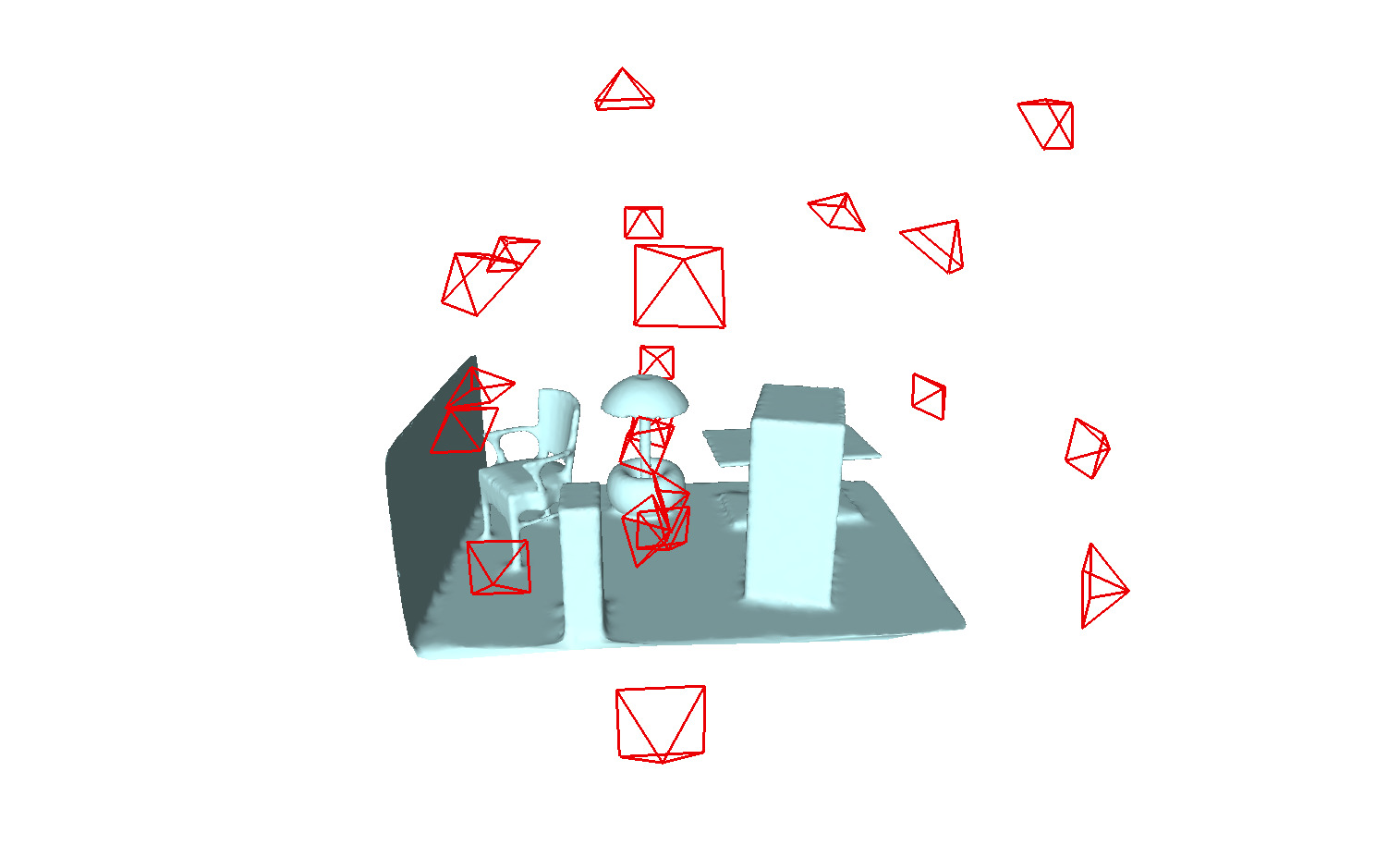}}
    \end{tabular}
    
    \caption{Virtual view generators. \textbf{Spherical Pattern Generator}: samples around the object while targeting it, always used for small objects. \textbf{Grid+Nadir Generator}: samples in grid fashion that parallel to the ground plane for the terrain-like large-scale scenario, uses above ground height and overlap rate to control the density of sampled views. \textbf{Grid+Oblique generator}: beside nadir views, samples concentric oblique views to capture facades in the urban area. \textbf{Customized}: we create an interactive tool to let users pick views. Customized views can improve the quality of ambiguous structures of complex objects. }
    \label{fig:viewgenerators}
\end{figure}

Given the input unstructured point cloud, we first generate the images under virtual views that simulate the input images in the MVS method. Virtual view sampling given a presumed object or scene is a challenging problem, and automatic view selection for tasks such as stereo reconstruction or surveillance is an active topic in computer vision and robotics community~\cite{nbv1999PAMI,smith2018aerial,zeng2020view,zengpc}, and the performance of these methods are task and scene-specific. In our work, we utilize several general generators as shown in Figure~\ref{fig:viewgenerators} and empirically find that they work very well with our method when dealing with different objects and scene contexts. In practice, our method shows a certain level of robustness regarding varieties of virtual view sampling, and the number of virtual views can affect surface quality, the analysis of which can be found in Section~\ref{sec:ablation}.
\subsection{Visibility Estimation Network\label{sec:method:networks}}
Given the projected image with sparse depth information, we exploit the deep neural network to predict the visibility of 3D points in the virtual views. The biggest challenge of the visibility estimation task is that it requires pixel-wise accuracy rather than region-wise. To this end, the standard CNNs (Convolutional Neural Networks) with small kernel size is preferred for the task because it is sensitive to local structure. However, for data with varying sparsity levels, the standard CNNs have poor performance since invalid pixels or voids produced by projecting sparse points make it difficult to learn robust representations~\cite{sparsityInvCnn}. 

Partial Convolution (PConv)~\cite{liu2018partialpadding, liu2018partialinpainting} is proposed for image inpainting from an incomplete input image. The proposed re-weighted convolution operation is particularly suitable for processing sparse data since it models the validity of each pixel and propagates it along with the convolutional operation. A similar idea has been used in depth completion~\cite{sparsityInvCnn} from sparse laser scan data. Visibility prediction exploits the local sensitivity of neural networks, while depth completion requires local smoothness. Therefore, we design an intermediate task with a dedicated network module for depth completion and proposed a cascade network VDVNet (Visibility-Depth-Visibility).


The input of our visibility estimation network is a $H \times W \times 2$ feature map in which the first channel is a normalized depth image, and the second is a binary mask that indicates whether the pixel is valid. Our frame shown in Figure~\ref{fig:architecture} includes an example input and intermediate results between different network components (which will be detailed in the following sections). The specifications of the network architecture can be found in our supplementary material.

\boldparagraph{CoarseVisNet.}
We train the sub-network to predict the visibility with supervised learning. As shown in Figure~\ref{fig:architecture}, we apply Sparse BCELoss (binary cross-entropy loss with mask) between the predicted coarse visibility map and the ground truth visibility map, where the mask indicates the valid pixels.

\boldparagraph{DepthCompNet.}
The depth completion sub-network is designed to convert a sparse depth map to a dense one. We filter raw depth map with coarse visibility predicted by CoarseVisNet before feeding it to DepthCompNet. The supervision of DepthCompNet is depth maps of ground truth surface. Mean squared error loss function is applied between them (Sparse MSELoss in Figure~\ref{fig:architecture}).

\boldparagraph{FineVisNet.}
This module takes the input of CoarseVisNet and the output of DepthCompNet to predict the fine visibility. The loss function is applied as same as CoarseVisNet.

\boldparagraph{Training Data.}
The training data used for visibility prediction can be easily simulated from any type of 3D model. We select a few well-reconstructed textured mesh models of large-scale scenes from a public MVS reconstruction dataset~\cite{yao2020blendedmvs} as the \textit{ground truth surfaces} to generate synthetic visibility. Points are uniformly and sparsely sampled from surfaces. Then, the sampled points are further augmented with additional Gaussian white noises and outliers. In the next step, we sample virtual views from the point clouds with the combination of patterns shown in Figure~\ref{fig:viewgenerators}. For each view, a depth map of both point clouds and surface are rendered and combined to generate the \textit{ground truth visibility}.
If the depth value of the point rendered image and the surface rendered image are matched, the point is labeled as visible, otherwise, as occluded. Depth maps of the surfaces serve as \textit{ground truth depth} for the training of DepthCompNet.

\subsection{Adaptive Visibility Weighting of Graph-Cut based Surface Reconstruction\label{sec:adaptiveweighting}}
Once the visibility of points in all views is predicted, we utilize the graph-cut based mesh generation~\cite{labatut2009robust} to reconstruct the final mesh model. Due to the different characteristics of virtual visibility, the basic method~\cite{labatut2009robust} suffers from the over-smoothing issue of creases on the surface, which motivates us to propose adaptive visibility weighting. The formulas of the basic method can be found in Section~\ref{sec:prelimiary}.

Compared with the visibility from instruments, the virtual view visibility generated by our VDVNet from unordered points and sampled views has very different characteristics in terms of density, direction, and distance.
As Equation~\ref{eq:origin_E_t} indicates, the basic method has an issue that the strong constraint imposed on the tetrahedra right behind the point along the line of sight and might overlook details of the surface especially when the lines of sight have a large incident angle. Since the large incident angle is rare in traditional applications, the issue of the baseline method was not discovered before.

\begin{figure}
    \centering
    \setlength\tabcolsep{4pt}
    \def\arraystretch{0}
    \begin{tabular}{cccc}
    \makecell{
    \subfloat[Sharp angle\label{fig:sharpanglesurface}]{
        \includegraphics[trim=3 35 3 5,clip,width=0.30\linewidth]{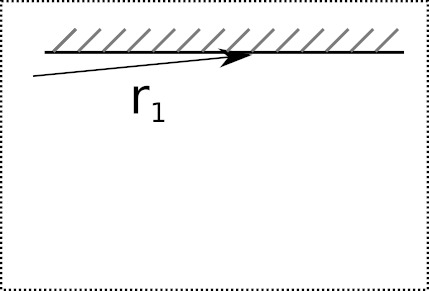}} \vspace{-1em}\\
        \subfloat[Fronto-parallel\label{fig:frontoparallelsurface}]{
        \includegraphics[trim=3 20 3 3,clip,width=0.30\linewidth]{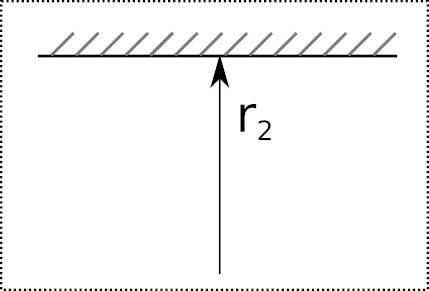}}
    }
     & 
     \makecell{\subfloat[$\gamma=0.1$\label{fig:avw_acute}]{
        \includegraphics[trim=8 15 10 5,clip,width=0.15\linewidth]{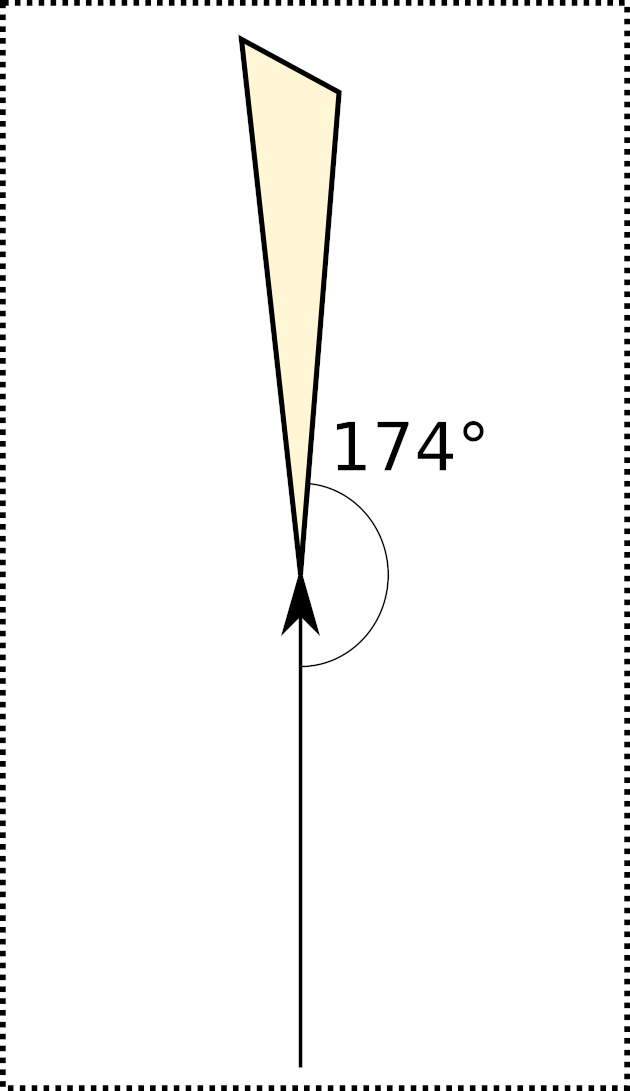}}}
    & \makecell{\subfloat[$\gamma=1$\label{fig:avw_right}]{
        \includegraphics[trim=8 15 10 5,clip,width=0.15\linewidth]{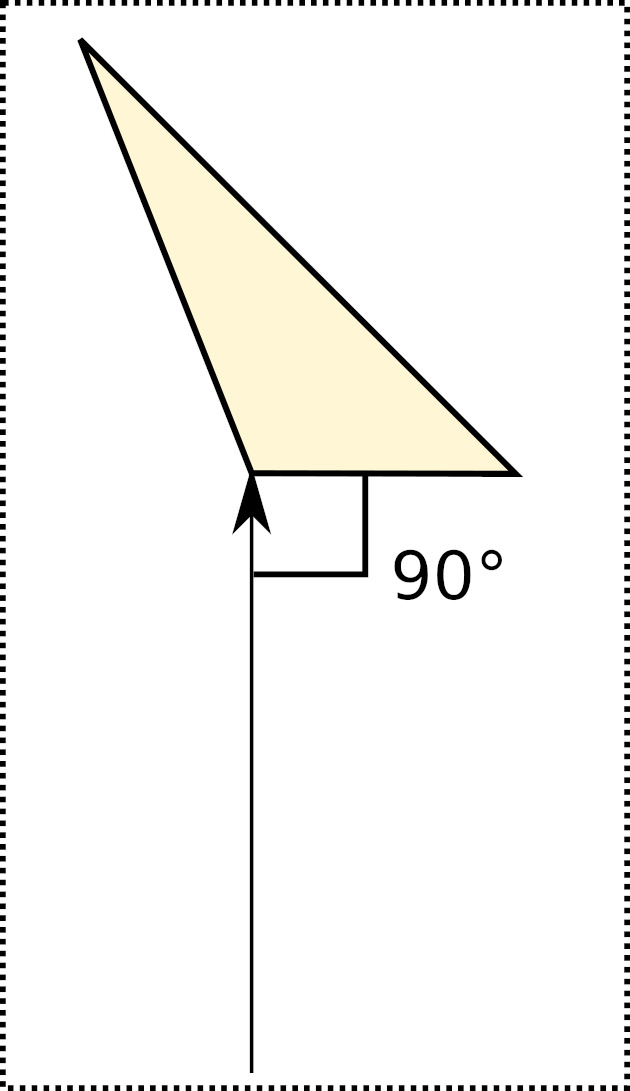}}}
    & \makecell{\subfloat[$\gamma=0.4$\label{fig:avw_obtuse}]{
        \includegraphics[trim=8 15 10 5,clip,width=0.15\linewidth]{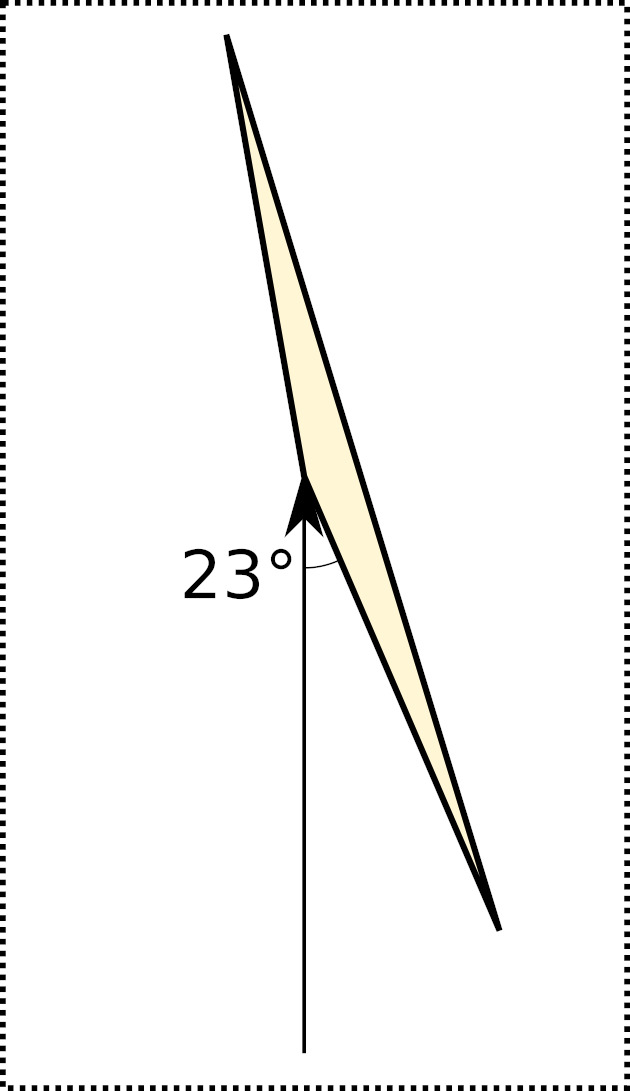}}}
    \end{tabular}
    
    \caption{Geometric illustration of adaptive visibility weighting, where $\lambda_{avw}=1$.}
    \label{fig:adaptive_vis_weighting}
\end{figure}

Since our virtual visibility is predicted by the neural network from unordered points directly, the angle of the incidence can range from 0 to 90 degrees, because there is no implicit filter of the multi-stereo view method. The visibility with a large incident angle (Figure~\ref{fig:sharpanglesurface}) is equivalent to a sharp surface in the depth map of the view, which is always considered as a sign of self-occlusion~\cite{bodis2015superpixelsharpangle}.

The key idea of our adaptive visibility weighting (AVW) is illustrated in Figure~\ref{fig:adaptive_vis_weighting}. Apparently, for surface reconstruction, a measurement for a sharp angle surface (Figure~\ref{fig:sharpanglesurface}) is considered as the lowest confidence, while for the fronto-parallel surface (Figure~\ref{fig:frontoparallelsurface}) is considered the best confidence. We use cosine similarity between the direction of the line of sight and the surface normal to construct a smooth weighting function. The weight is regarded as the confidence of visibility, its definition overlaps with $\alpha_{vis}$, and it changes terms $E_t$ and $E_{ij}$. We use the product of two factors $\gamma$ and $\alpha_{vis}$ as the new confidence of line of sight.
In order to apply the weighting function to a tetrahedron with 3 triangles where the vertex and the line of sight intersect, we consider the one most likely to be on the surface, that is, the triangle whose normal is closest to the direction of the ray. We took the maximum value of the cosine similarity of the three angles, as shown in Equation~\ref{eq:our_energy}, where the cosine similarity is equivalent to the dot product because the normal is of unit length. Figure~\ref{fig:avw_acute}, \ref{fig:avw_right}, \ref{fig:avw_obtuse} show examples of our adaptive visibility weighting term. 
\begin{subequations}
\small
\vspace{-0.2em}
\begin{align}
& E^{OURS}_{t}(v) = \gamma \cdot \alpha_{vis} \cdot \delta[T_{M+1} \in s] \label{eq:our_E_t},\\
& E^{OURS}_{ij}(v) = \sum_{i=1}^{M-1} \gamma \cdot \alpha_{vis} \cdot \delta[T_i \in s \wedge T_{i+1} \in t] \label{eq:our_E_ij}, \\
& \gamma = (1-\lambda_{avw})+\lambda_{avw}\max(N_v^T\cdot [N_{f1}, N_{f2}, N_{f3}]), \label{eq:ourAVW}
\end{align}
\label{eq:our_energy}
\end{subequations}
where $N_v$ denotes direction of the incident line of sight $v$, $N_{f\#}$ are normal of faces of tetrahedron $T_{M+1}$ incident to the endpoint of $v$, $\lambda_{avw}$ is the damping factor to adjust the adaptive visibility weighting $\gamma$. Our method is equivalent to Labatut's method~\cite{labatut2009robust} when $\lambda_{avw}=0$.

\section{Experiments}
We evaluate our proposed Vis2Mesh method through a series of quantitative and qualitative experiments covering a variety of small objects, indoor scenarios, large-scale outdoor scenarios either acquired by multi-view stereo, laser scanning, or simulated images. We compare our method with several state-of-the-art learning-based reconstruction approaches including Point2Mesh (P2M)~\cite{Hanocka2020p2m}, Points2Surf (P2S)~\cite{ErlerEtAl:Points2Surf:ECCV:2020}, Meshing Point Cloud with IER (MIER)~\cite{liu2020meshing} Convolutional Occupancy Networks(CONet)~\cite{Peng2020ECCV}, and a classical approach Screened Poisson Surface Reconstruction (SPSR)~\cite{kazhdan2013screened}. To demonstrate the generalization capacity, we widely collect data from multiple datasets with varying sensors and platforms, including BlendedMVS~\cite{yao2020blendedmvs}, COSEG~\cite{van2011prior}, Thingi10K~\cite{zhou2016thingi10k}, P2M~\cite{Hanocka2020p2m}, CONet~\cite{Peng2020ECCV}, senseFly~\cite{sensefly}, KITTI~\cite{kitti2013}, Mai City Dataset~\cite{maidataset}, Airborne LiDAR data of Columbus, Ohio~\cite{OGRIP} and Airborne LiDAR data of Toronto~\cite{rottensteiner2012isprs}. 
Details of these datasets including their source, volume, as well as virtual view generators used in our experiments can be found in the supp. material.

Our method does not assume surface normal information, while it is required for some of the comparing methods (SPSR, P2M, \etc). High-quality surface normals are an important source of information to guide the reconstruction method but are not easily obtainable or directly measured by most of the sensors. In the experiment, the surface normal used by these comparing methods is estimated by a Minimal Spanning Tree (MST) based local plane fitting method, implemented by the open-source software CloudCompare~\cite{cloudcompare}. The software also integrated SPSR~\cite{kazhdan2013screened} implementation which we use in our comparative study. We report the performance of the comparing methods quantitatively in Section~\ref{sec:quant}, and qualitatively in Section~\ref{sec:quali}.  

\subsection{Quantitative Evaluation\label{sec:quant}}

\begin{table*}
    \centering
    \resizebox{\linewidth}{!}{
    \begin{tabular}{|c|c|c|c|c|c|c|c|c|c|c|c|c|c|c|c|}
    \hline
    \multicolumn{2}{|c|}{} & \multicolumn{7}{c|}{\textbf{F-score}} & \multicolumn{7}{c|}{\textbf{Chamfer distance}} \\\cline{3-16}

    \multicolumn{2}{|c|}{} & \multirow{2}{*}{P2M} & \multirow{2}{*}{P2S} & \multirow{2}{*}{MIER} & \multirow{2}{*}{CONet} & SPSR & Ours & Ours & \multirow{2}{*}{P2M} & \multirow{2}{*}{P2S} & \multirow{2}{*}{MIER} & \multirow{2}{*}{CONet} & SPSR & Ours & Ours \\
    \multicolumn{2}{|c|}{} & &  &  &  & (Trimmed) & ($\gamma_{av}=0$) & ($\gamma_{av}=1$) &  &  &  &  & (Trimmed) & ($\gamma_{av}=0$) & ($\gamma_{av}=1$) \\
    \hline\hline
    \multirow{4}{*}{\makecell{Object\\25K}} & DSLR & 0.9790 & 0.9014 & \textbf{0.9935} & 0.9041 & 0.9517 & 0.9844 & 0.9841 & 0.1593 & 0.3558 & 0.1440 & 0.2570 & 0.1288 & \textbf{0.0702} & 0.0707\\
    & bull & 0.9990 & 0.9946 & 0.9851 & 0.8203 & \textbf{0.9993} & 0.9929 & 0.9991 & 0.3328 & 0.3576 & 0.3421 & 1.3881 & 0.2914 & 0.3037 & \textbf{0.2893} \\
    & giraffe & 0.9969 & 0.9525 & 0.9938 & 0.9776 & \textbf{0.9999} & 0.9989 & 0.9996 & 0.2227 & 0.4325 & 0.3276 & 0.3831 & \textbf{0.1893} & 0.2136 & 0.1939\\ \cline{2-16}
    & \textit{Average} & \textit{0.9916} & \textit{0.9495} & \textit{0.9908} & \textit{0.9007} & \textit{0.9836} & \textit{0.9921} & \textit{\textbf{0.9943}} & \textit{0.2383 }& \textit{0.3819} & \textit{0.2712} & \textit{0.6761} & \textit{0.2032 }& \textit{0.1959} & \textit{\textbf{0.1846}}\\
    \hline
    \multirow{3}{*}{\makecell{Indoor\\100K}}& room0 & / & 0.7334 & 0.8885 & 0.9112 & 0.9049 & \textbf{0.9345} & 0.9340 & / & 2.0151 & 0.3403 & 0.6476 & 0.8272 & \textbf{0.3296} & 0.3571\\
    & room1 & / & 0.5379 & 0.8453 & 0.8695 & 0.8065 & 0.8747 & \textbf{0.8775} & / & 2.8026 & 0.6714 & 0.7814 & 1.2546 & 0.5971 & \textbf{0.5457}\\\cline{2-16}
    & \textit{Average} & / & \textit{0.6356} & \textit{0.8669} & \textit{
0.8903} & \textit{0.8557} & \textit{0.9046} & \textit{\textbf{0.9058}} & / & \textit{2.4088} & \textit{0.5059} & \textit{0.7145} & \textit{1.0409} & \textit{0.4634} & \textit{\textbf{0.4514}}\\
    \hline
    \multirow{6}{*}{\makecell{Outdoor\\500K}}& Church & / & / & / & 0.4086 & 0.8292 & 0.9206 & \textbf{0.9221} & / & / & / & 19.8253 & 1.7784 & 0.6811 & \textbf{0.6657}\\
    & Archway & / & / & / & 0.9133 & 0.9815 & \textbf{0.9840} & 0.9772 & / & / & / & 0.2753 & 0.0869 & \textbf{0.0677} & 0.0756\\
    & Pedestrian street & / & / & / & 0.7494 & 0.9401 & \textbf{0.9889} & 0.9860 & / & / & / & 0.8008 & 0.3561 & \textbf{0.1590} & 0.1620\\
    & Eco Park & / & / & / & 0.7416 & 0.8470 & 0.9431 & \textbf{0.9553} & / & / & / & 0.8813 & 0.9463 & 0.3273 & \textbf{0.2961}\\
    & Dragon Park & / & / & / & 0.7700 & 0.8702 & 0.9804 & \textbf{0.9866} & / & / & / & 0.8696 & 0.7398 & 0.2096 & \textbf{0.1859}\\\cline{2-16}
    & \textit{Average} & / & / & / & \textit{0.7166} & \textit{0.8936} & \textit{0.9634} & \textit{\textbf{0.9654}} & / & / & / & \textit{4.4809} & \textit{0.7815} & \textit{0.2890} & \textit{\textbf{0.2770}}\\
    \hline
    \multirow{4}{*}{MVS} & Hotel & / & / & / & 0.3577 & 0.7732 & 0.8229& \textbf{0.8350} & / & / & / & 1.3053 &	0.9369 & 0.3660 & \textbf{0.3479} \\
    & GSM Tower & / & / & / & 0.1863 & 0.6261 & 0.6321 & \textbf{0.6354} & / & / & / & 6.6513 & 0.8051 & 0.6935 & \textbf{0.6888} \\
    & UThammasat & / & / & / & 0.4656 & \textbf{0.8046} & 0.7094 & 0.7259 & / & / & / & 1.4951  & 0.3741 & 0.3982 & \textbf{0.3737} \\\cline{2-16}
    & \textit{Average} & / & / & / & \textit{0.3365} & \textit{\textbf{0.7346}} & \textit{0.7215} & \textit{0.7321} & / & / & / & \textit{1.7845} & \textit{0.7054} & \textit{0.4859} & \textit{\textbf{0.4701}} \\
    \hline
    \end{tabular}
    }
    \caption{Quantitative comparison of reconstruction on COSEG, CONet, and BlendedMVS dataset, grouped by the size of the scene and number of points. (``F-score": Higher is better. ``Chamfer distance": Lower is better. ``/": Not applicable.)}
    \label{tab:quantitative_detail}
\end{table*}

We evaluate our method on points sampled from ground truth mesh from COSEG, CONet, and BlendedMVS datasets. Since the existing methods have certain limitations on the amount of data, we divide the data set into object/indoor/outdoor scenes, and the complexity and scene scale are gradually increasing. We uniformly sample 25K points from the ground truth surface for objects, 100K points for indoor data, and 500K points for outdoor data. 
In order to evaluated with actual data, we generated ground truth for the sensefly~\cite{sensefly} dataset by using a commercial MVS pipeline~\cite{pix4d} that employs SPSR for surface reconstruction. Table~\ref{tab:quantitative_detail} shows F-score~\cite{tanksandtemples} and Chamfer distance (CD)~\cite{barrow1977parametric} metrics on each category. To be noted that, we use the sliding window manner of CONet for all scales, and we trim SPSR output by the density value, and the threshold is set to $max\_density/2$.

Compared with several state-of-the-art learning-based reconstruction approaches and the SPSR method, our virtual view visibility-based methods present outstanding performance in terms of both metrics on each level of datasets, which also proves the insensitivity of our method to the type of scene.
Furthermore, although our VDVNet is trained from the synthetic dataset collected from outdoor-level models, and the quantitative evaluation shows our method has achieved competitive results on object-level and indoor-level datasets. 
Regarding efficiency, our method has similar speed and memory as SPSR when reaching the same level of detail. Detailed statistics on running time can be found in the supp. material.

\subsection{Ablation Studies\label{sec:ablation}}
\begin{figure}
    \captionsetup[subfigure]{justification=centering}
    \def\subimgwidth{0.27\linewidth}
    \centering
    \subfloat[Points]{%
    \begin{overpic}[width=\subimgwidth]{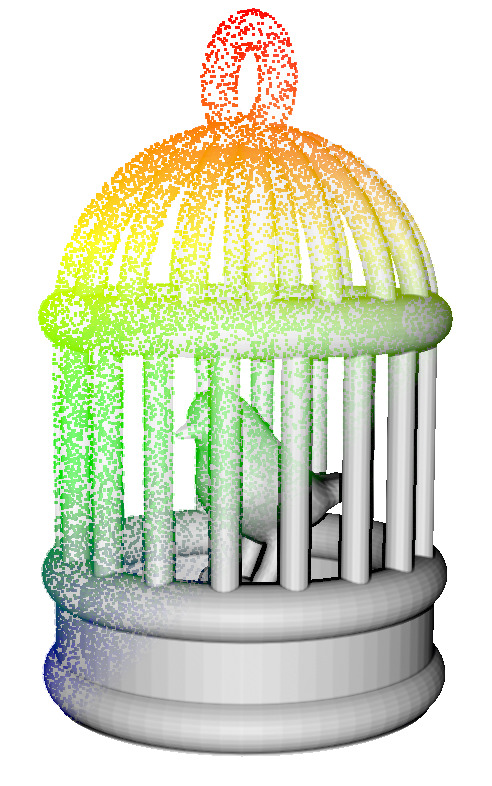}
    \end{overpic}
    }
    \subfloat[$\lambda_{avw}=0$]{%
    \begin{overpic}[width=\subimgwidth]{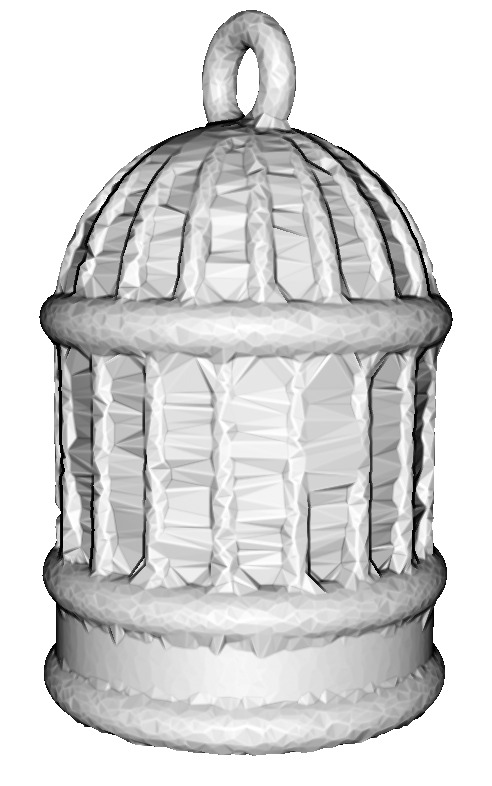}
    \put(0,95){\scriptsize F-score:}
    \put(10,90){\scriptsize 0.64}
    \put(42,95){\scriptsize CD:}
    \put(42,90){\scriptsize 1.80}
    \end{overpic}
    }
    \subfloat[$\lambda_{avw}=1$]{%
    \begin{overpic}[width=\subimgwidth]{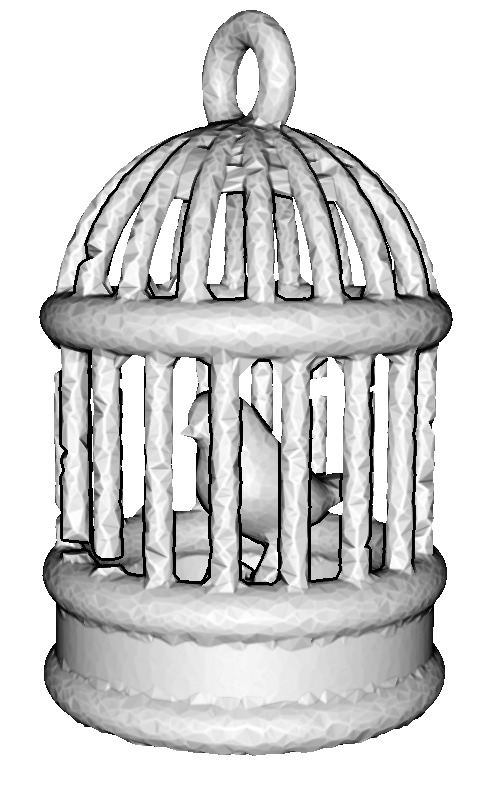}
    \put(0,95){\scriptsize F-score:}
    \put(10,90){\scriptsize 0.97}
    \put(42,95){\scriptsize CD:}
    \put(42,90){\scriptsize 0.23}
    \end{overpic}
    }
    \caption{Impact of AVW on synthetic data with complex shape. The data is \textit{Birdcage}.}
    \label{fig:ablation:angular_small}
\end{figure}
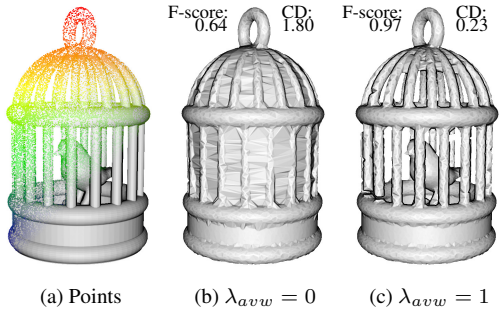

We first demonstrate the advantage of virtual view visibility in surface reconstruction even if the actual visibility is available (Figure~\ref{fig:importance_of_vvv}), and show the quality improvement by the proposed adaptive visibility weighting (Figure~\ref{fig:ablation:angular_small}). Then, we analyze the network architecture with a quantitative evaluation (Table~\ref{tab:network_quantitative}).

\begin{figure}
    \def\subimgwidth{0.32\linewidth}
    \centering
    \subfloat[Per Point]{
        \includegraphics[width=\subimgwidth,trim=15 0 30 0,clip]{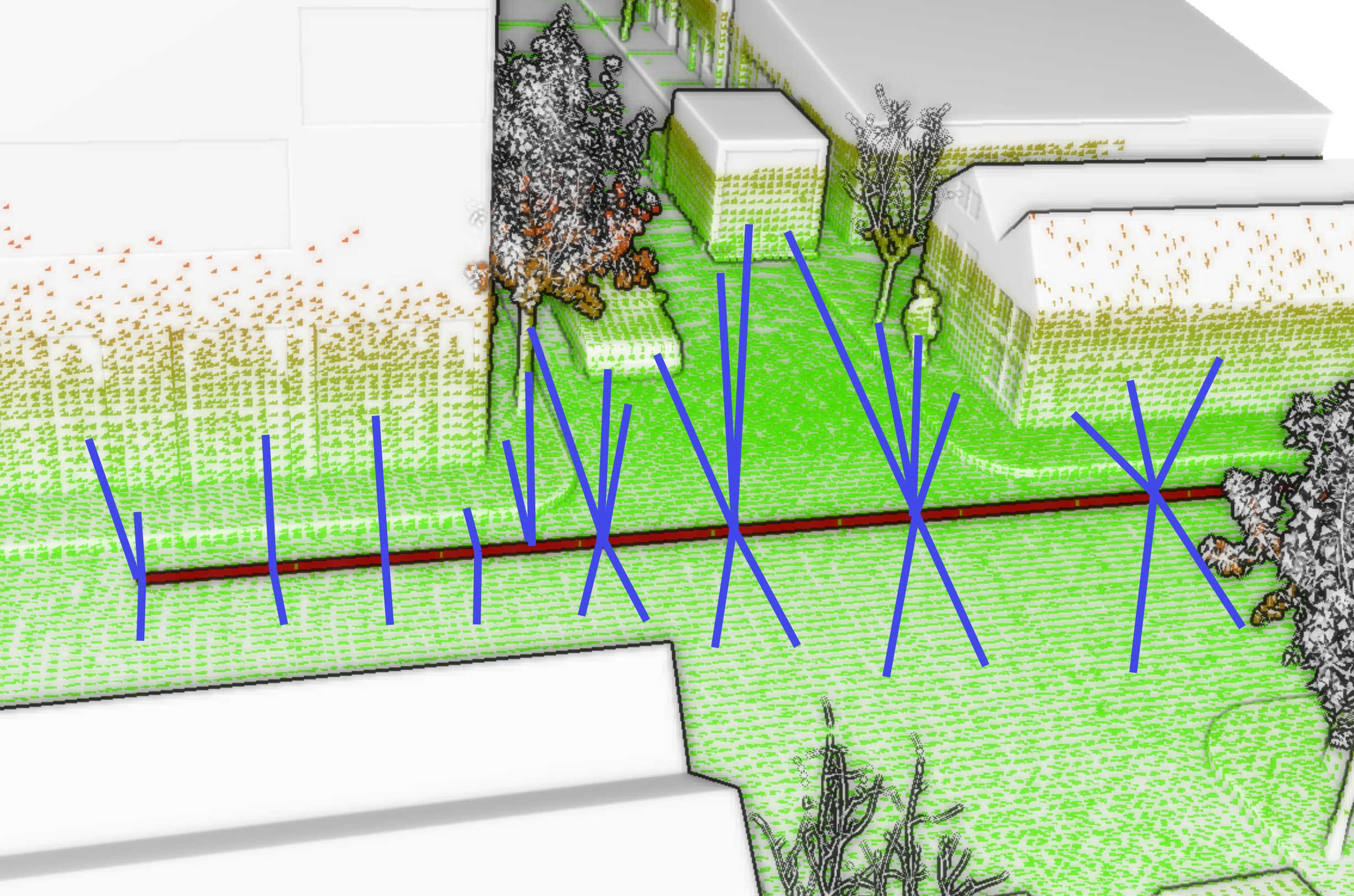}
    }
     \subfloat[Grid Fusion]{
        \includegraphics[width=\subimgwidth,trim=15 0 30 0,clip]{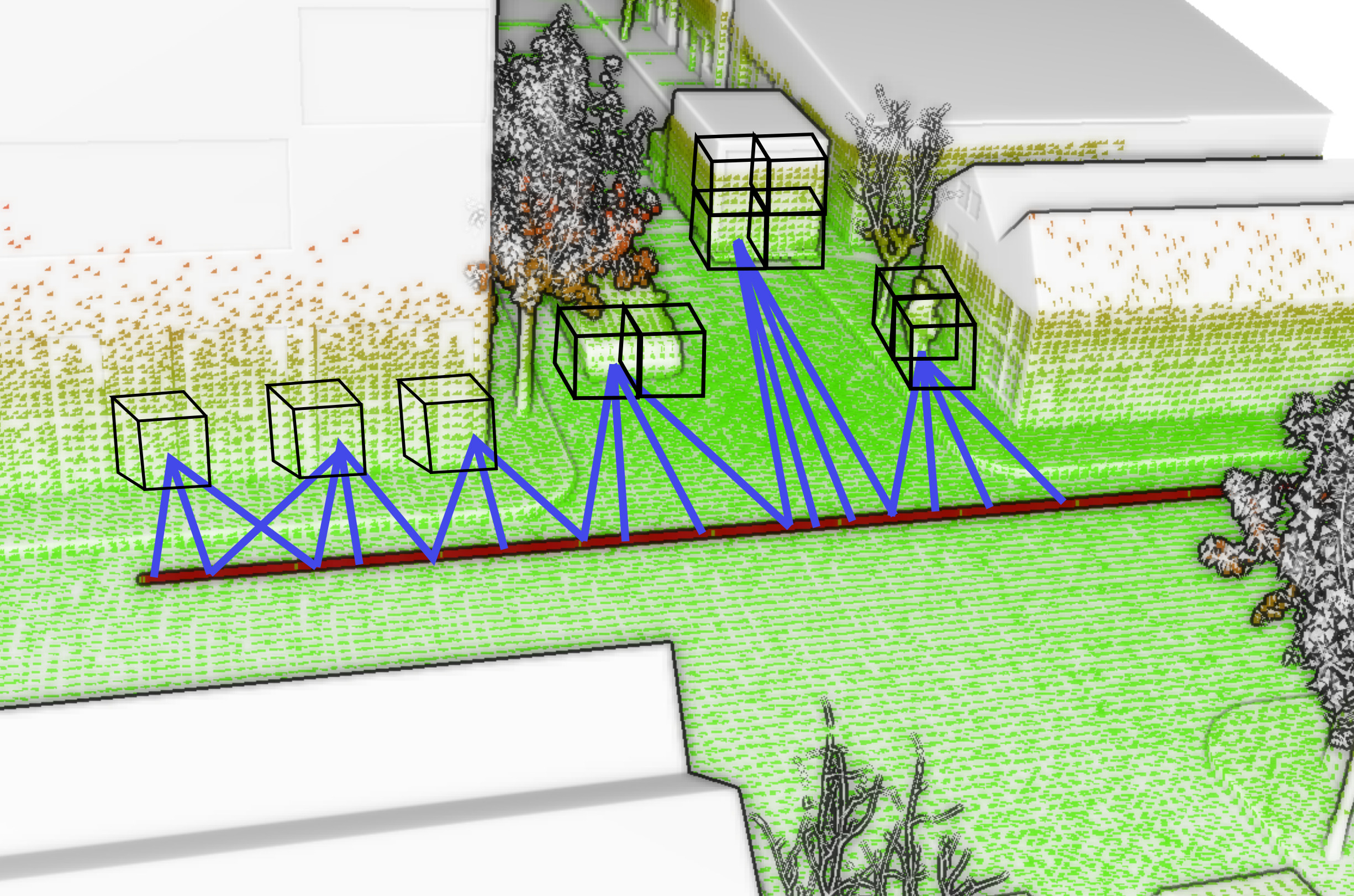}
    }
     \subfloat[Ours]{
        \includegraphics[width=\subimgwidth,trim=15 0 30 0,clip]{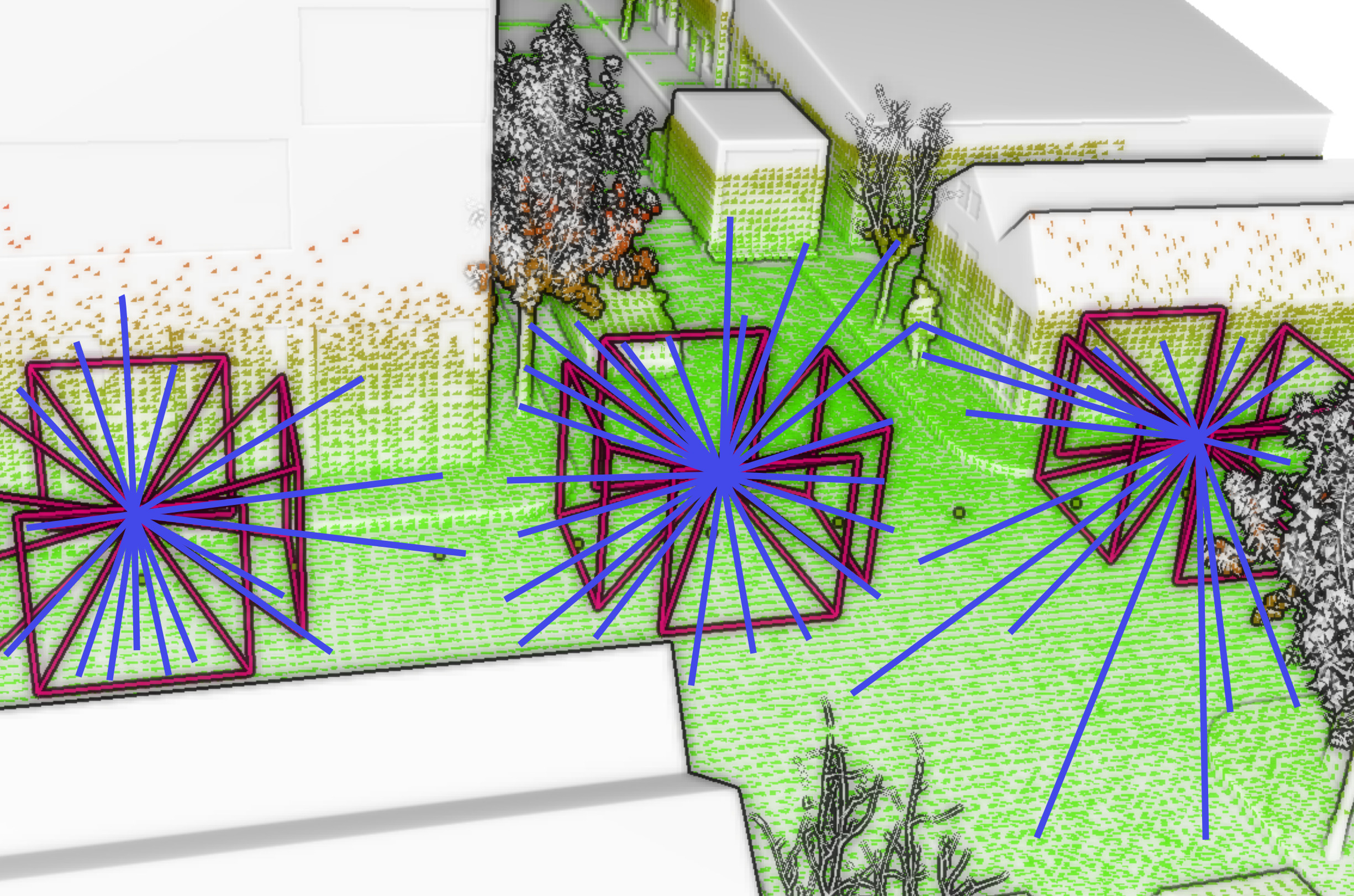}
    }
    \caption{Methods of modeling visibility from mobile LiDAR scans. (Per Point: each LiDAR point only connects to the trajectory once. Grid Fusion: points within the voxel of the grid fuse into a centroid and connect it to the trajectory. Ours: pure virtual visibility generated from sampled views and VDVNet.) }
    \label{fig:kitti_visiblity_source}
\end{figure}

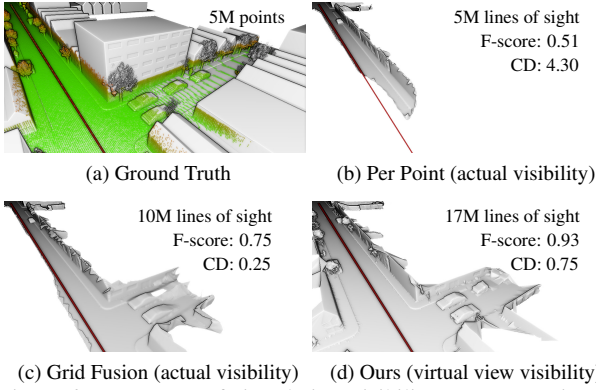
\begin{figure}
    \centering
    \subfloat[Ground Truth]{%
    \begin{overpic}[width=0.48\linewidth,trim=50 100 10 20,clip]{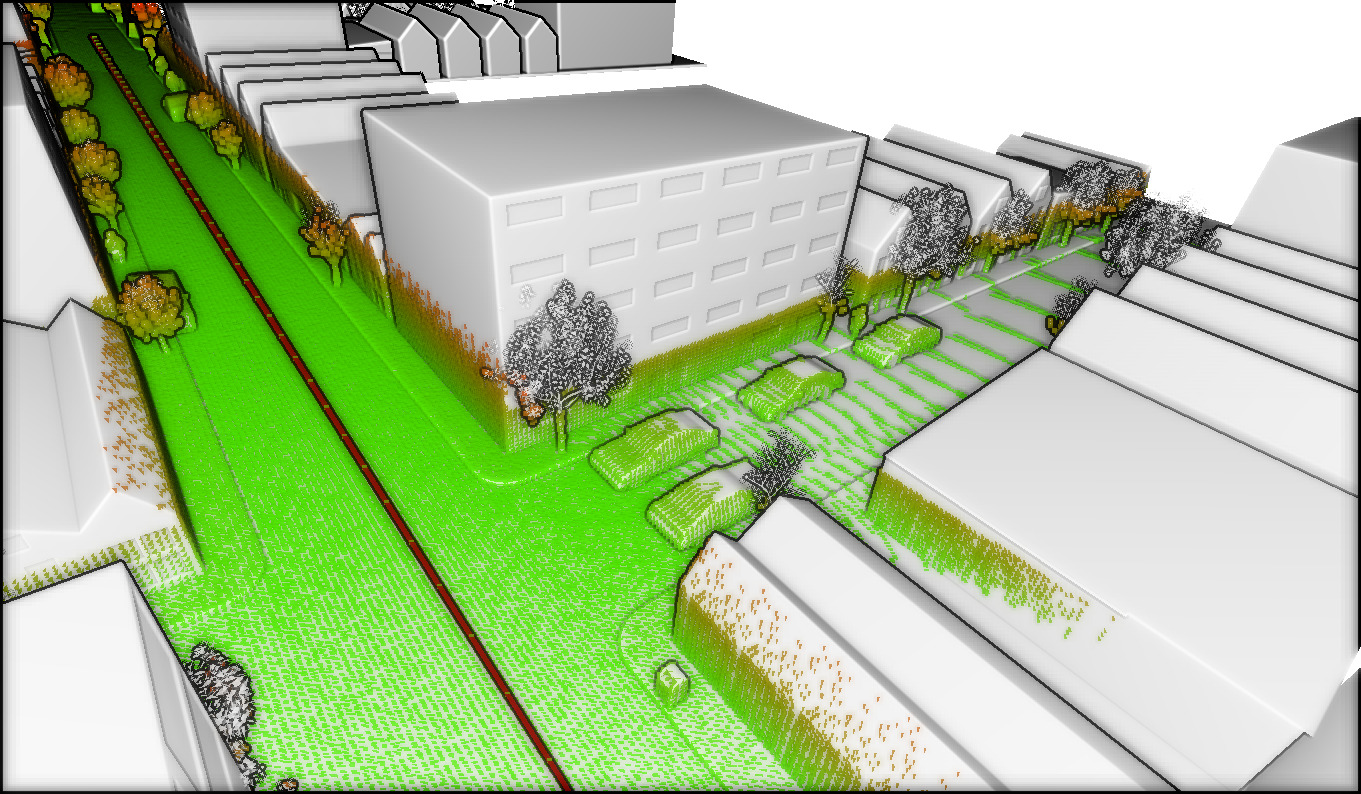}
    \put(68,43){\scriptsize 5M points}
    \end{overpic}
    }\subfloat[Per Point (actual visibility)]{%
    \begin{overpic}[width=0.48\linewidth,trim=50 100 10 20,clip]{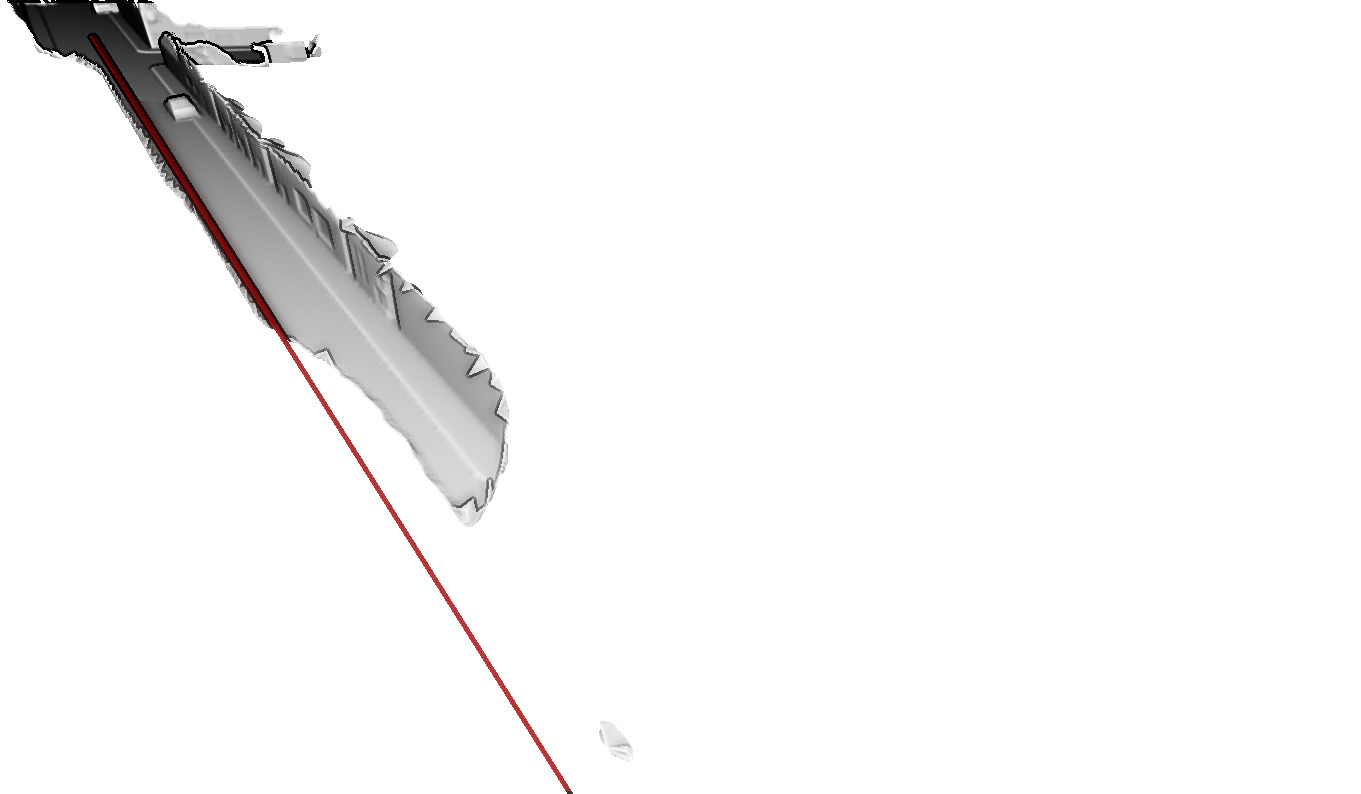}
    \put(47,43){\scriptsize 5M lines of sight}
    \put(56,35){\scriptsize F-score: 0.51}
    \put(66,27){\scriptsize CD: 4.30}
    \end{overpic}
    }\\\vspace{-0.5em}
    \subfloat[Grid Fusion (actual visibility)]{%
    \begin{overpic}[width=0.48\linewidth,trim=50 100 10 20,clip]{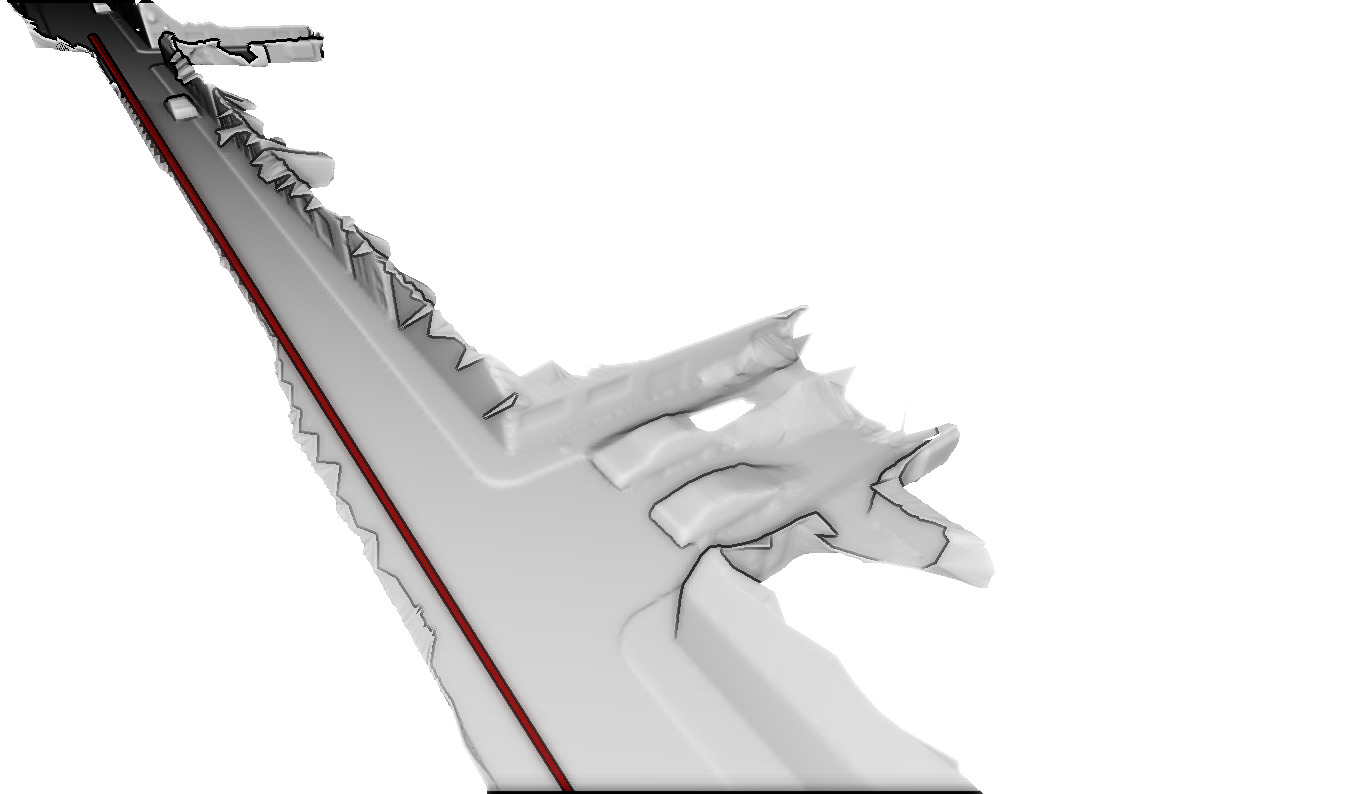}
    \put(44,43){\scriptsize 10M lines of sight}
    \put(56,35){\scriptsize F-score: 0.75}
    \put(66,27){\scriptsize CD: 0.25}
    \end{overpic}
    }
    \subfloat[Ours (virtual view visibility)]{%
    \begin{overpic}[width=0.48\linewidth,trim=50 100 10 20,clip]{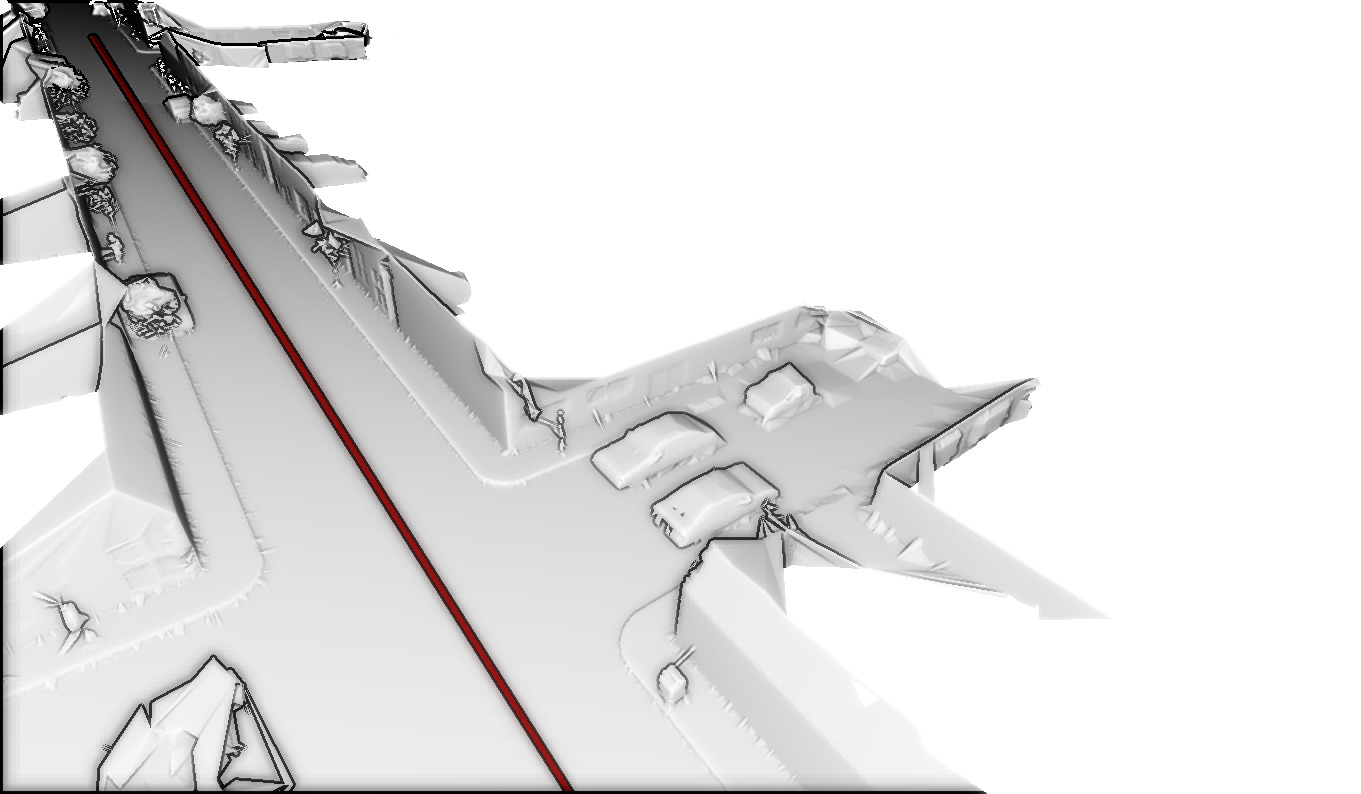}
    \put(44,43){\scriptsize 17M lines of sight}
    \put(56,35){\scriptsize F-score: 0.93}
    \put(66,27){\scriptsize CD: 0.75}
    \end{overpic}
    }
    
    \caption{Importance of virtual view visibility. Reconstruction using our method for each visibility dataset in Figure~\ref{fig:kitti_visiblity_source}.}
    \label{fig:importance_of_vvv}
\end{figure}

\boldparagraph{Importance of virtual view visibility.}
In this experiment, we compare surfaces reconstructed by real visibility (obtained from physical views) and virtual view visibility to prove the importance of virtual visibility. The data in this experiment is LiDAR scans associated with 6DoF pose trajectory from the Mai City Dataset~\cite{kitti2013}. We present three methods as shown in Figure~\ref{fig:kitti_visiblity_source}. \textit{Per Point} is generated by associating the corresponding scan pose with each point, and each point has only one line of sight (or visibility) collected from the sensor.
For \textit{Grid Fusion}, we use a voxel grid filter to aggregate nearby points' visibility to obtain points with multiple actual visibility. Finally, we use the proposed network to generate virtual view visibility, where the virtual views are sampled along the recorded trajectory. 

The visual results shown in Figure~\ref{fig:importance_of_vvv} demonstrates that the virtual views visibility place a significant improvement in the mesh quality. Per Point actual visibility failed to reconstruct a complete scene because only one line of sight associated with each point is not geometrical stable, and is not sufficient to create a reliable weighted graph for the optimization. Grid Fusion shows better reconstruction quality since more lines of sight are included  (4/points on average), but the details of reconstruction are poor, and noises seriously affect the surface. In contrast, our pure virtual view visibility method creates smooth, detailed, and noise-free surfaces (thanks to the denoising capability of VDVNet).

\boldparagraph{Importance of adaptive visibility weighting.}
Figure~\ref{fig:ablation:angular_small} demonstrates the impact of proposed adaptive visibility weighting. To be noted that when setting $\lambda_{avw}=0$, the visibility weighted graph is exactly equivalent to the basic method~\cite{labatut2009robust}. We observe the method $\lambda_{avw}=1$ works better than the basic method for objects with pole-like and notches on the surface structure, since such a complex structure may introduce large incidence angles for our virtual lines of sight. The adaptive visibility weighting works as we expected in Section~\ref{sec:adaptiveweighting}. The quantitative results in Table~\ref{tab:quantitative_detail} show that in most cases, adaptive visibility weighting works better than the basic method.


\begin{figure}
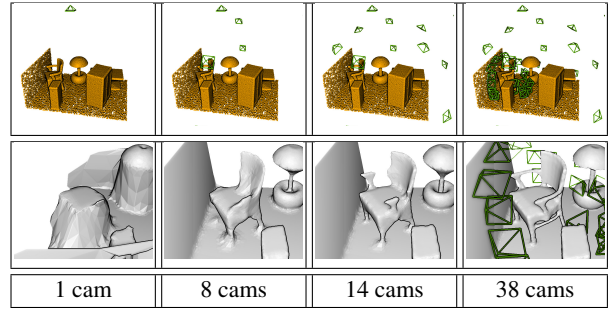

    \def\subimgwidth{0.22\linewidth}
    \edef\trimopta{trim=270 50 400 110,clip,width = \subimgwidth}
    \edef\trimoptb{trim=120 50 140 145,clip,width = \subimgwidth}
    \edef\trimoptc{trim=60 100 620 90,clip,width = \subimgwidth}
    
    \centering
    \setlength\tabcolsep{1pt}
    \begin{tabular}{|c||c||c||c|}
        \hline
        \expandafter\includegraphics\expandafter[\trimopta]{room_0_viewanalysis/orangecloud_c0_v1.jpg} &
        \expandafter\includegraphics\expandafter[\trimopta]{room_0_viewanalysis/orangecloud_c1_v1.jpg} &
        \expandafter\includegraphics\expandafter[\trimopta]{room_0_viewanalysis/orangecloud_c2_v1.jpg} &
        \expandafter\includegraphics\expandafter[\trimopta]{room_0_viewanalysis/orangecloud_c4_v1.jpg} \\\hline\hline
        \expandafter\includegraphics\expandafter[\trimoptc]{room_0_viewanalysis/our_c0_v2.jpg} &
        \expandafter\includegraphics\expandafter[\trimoptc]{room_0_viewanalysis/our_c1_v2.jpg} &
        \expandafter\includegraphics\expandafter[\trimoptc]{room_0_viewanalysis/our_c2_v2.jpg} &
        \expandafter\includegraphics\expandafter[\trimoptc]{room_0_viewanalysis/our_c4_v2.jpg}
        \\\hline\hline
        \small 1 cam & \small 8 cams &
        \small 14 cams &
        \small 38 cams \\\hline
    \end{tabular}
    \caption{Impact of the number of virtual views.}
    \label{fig:qual:viewconfig}
\end{figure}

\boldparagraph{Impact of the number of virtual views.} 
To evaluate the effect of the view configuration, we demonstrate the reconstruction results \wrt different numbers of virtual views in Figure~\ref{fig:qual:viewconfig}. Firstly, the single view on the top center of the scene generates a very coarse surface. Secondly, 7 virtual views on the spherical orbit around the target carved out objects of the scene, the base of the lamp, legs of the chair starting emerging. With more virtual views, the reconstructed surface yields more geometric details. The last configuration contains virtual views selected interactively by the user to reconstruct the complex structure of the chair.

\boldparagraph{Visibility estimation networks.}
We independently evaluate the performance of our VDVNet in the task of visibility prediction comparing with geometric-based method Hidden Point Removal (HPR)~\cite{katz2007direct} implemented by Open3D~\cite{open3d} and the baseline learning method UNet~\cite{unet2015}. 
The \textsc{VisibNet} proposed by InvSFM~\cite{pittaluga2019revealing} is designed for the same task as us, and its architecture is UNet with convolutional layers at the end of the decoder.

We report the size and the performance of classifiers in Table~\ref{tab:network_quantitative} on our pixel-wise visibility dataset sampled from BlendedMVS. We observe the networks with partial convolutional layers outperform standard convolutional networks without parameter increasing. The variants of our networks present the contribution of the depth completion module and partial convolutional layers \wrt visibility estimation task. 
Please refer to the supp. material for the evaluation ofdatasets with various modalities as well as the qualitative results of visibility estimation.
\begin{table}[tp]
    \centering
    \resizebox{\linewidth}{!}{
        \begin{tabular}{|c|c|c|c|c|c|}
        \hline
        \textbf{Visibility Estimator} & \#Param & \%P & \%R & \%F1 & \%AUC \\
        \hline\hline
        HPR & / & 85.79 & 85.01 & 85.24 & 82.22\\ \hline
        UNet & 51M & 90.42 & 87.10 & 88.68 & 88.52 \\
        UNet + PConv & 51M & 90.73 & 88.03 & 89.32 & 89.59\\ 
        \textsc{VisibNet} & 52M & 90.15 & 91.50 & 90.78 & 90.22 \\ \hline
        Ours w/o DepthComp or PConv & 153M & 91.84 & 84.08 & 87.71 & 89.29\\
        Ours w/o DepthComp & 153M & 91.38 & 93.10 & 92.21 & 92.62\\
        Ours w/o PConv & 153M & 91.86 & 94.24 & 93.01 & 93.14\\
        Ours & 153M & \textbf{92.37} & \textbf{94.95} & \textbf{93.63} & \textbf{94.17} \\
        \hline
        \end{tabular}
        }
    \caption{Quantitative analysis of methods on binary visibility classification task. \#Param is the number of parameters, \%P is precision, \%R is recall, \%F1 is F1 score, and \%AUC is Area Under The Curve of ROC curves. DepthComp represents the depth completion module and PConv denotes t partial convolutional layers.}
    \label{tab:network_quantitative}
    \vspace{-0.8em}
\end{table}


\boldparagraph{Impact of noises and missing data.}
Since the real 3D point clouds normally have noises and incomplete data, it is very important for mesh generation to deal with these problems. We designed two experiments to evaluate the robustness of different methods against the additional noises and incompleteness.  

\begin{figure}
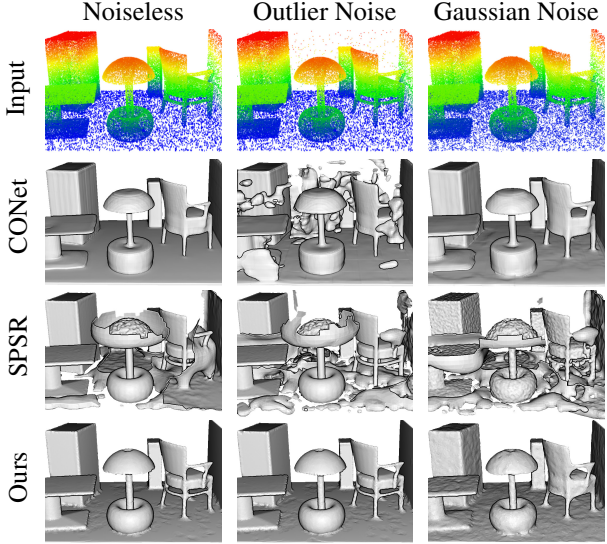

    \def\subimgwidth{0.28\linewidth}
    \edef\trimopt{trim=350 0 230 165,clip,clip,width = \subimgwidth}
    
    \centering
    \setlength\tabcolsep{3pt}
    \def\arraystretch{1}
    \begin{tabular}{cccc}
        & Noiseless & Outlier Noise & Gaussian Noise \\
        \rotatebox[origin=c]{90}{Input} &
        \raisebox{-0.5\height}{\expandafter\includegraphics\expandafter[\trimopt]{indoor_room_0_raw/pointcloud_ramp_view_1.jpg}} &
        \raisebox{-0.5\height}{\expandafter\includegraphics\expandafter[\trimopt]{indoor_room_0_outlier/pointcloud_ramp_view_1.jpg}} &
        \raisebox{-0.5\height}{\expandafter\includegraphics\expandafter[\trimopt]{indoor_room_0_noise/pointcloud_ramp_view_1.jpg}} \\
        \rotatebox[origin=c]{90}{CONet} &
        \raisebox{-0.5\height}{\expandafter\includegraphics\expandafter[\trimopt]{indoor_room_0_raw/convonet_view_1.jpg}} &
        \raisebox{-0.5\height}{\expandafter\includegraphics\expandafter[\trimopt]{indoor_room_0_outlier/convonet_view_1.jpg}} &
        \raisebox{-0.5\height}{\expandafter\includegraphics\expandafter[\trimopt]{indoor_room_0_noise/convonet_view_1.jpg}} \\
        \rotatebox[origin=c]{90}{SPSR} &
        \raisebox{-0.5\height}{\expandafter\includegraphics\expandafter[\trimopt]{indoor_room_0_raw/poisson_trim_view_1.jpg}} &
        \raisebox{-0.5\height}{\expandafter\includegraphics\expandafter[\trimopt]{indoor_room_0_outlier/poisson_trim_view_1.jpg}} &
        \raisebox{-0.5\height}{\expandafter\includegraphics\expandafter[\trimopt]{indoor_room_0_noise/poisson_trim_view_1.jpg}} \\
        \rotatebox[origin=c]{90}{Ours} &
        \raisebox{-0.5\height}{\expandafter\includegraphics\expandafter[\trimopt]{indoor_room_0_raw/ours_view_1.jpg}} &
        \raisebox{-0.5\height}{\expandafter\includegraphics\expandafter[\trimopt]{indoor_room_0_outlier/ours_view_1.jpg}} &
        \raisebox{-0.5\height}{\expandafter\includegraphics\expandafter[\trimopt]{indoor_room_0_noise/ours_view_1.jpg}}
        
    \end{tabular}
    \caption{Evaluation of the methods subject to additional noises on the point clouds. The data is \textit{Room0}.}
    \label{fig:noise:additional}
\end{figure}

\begin{figure}
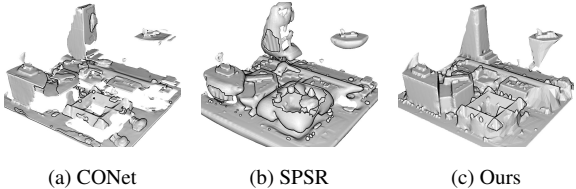

    \def\subimgwidth{0.28\linewidth}
    \edef\trtCopts{trim=190 60 380 50,clip,width = \subimgwidth}
    \edef\trtDopts{trim=200 80 300 70,clip,width = \subimgwidth}
    \edef\clbBopts{trim=0 0 0 20,clip,width = \subimgwidth}
    
    \centering
    \setlength{\tabcolsep}{2pt}
    \def\arraystretch{1}
    \begin{tabular}{ccc} 
    \subfloat[CONet]{%
    \begin{tabular}{c}
         \expandafter\includegraphics\expandafter[\trtCopts]{largescale_toronto/convonet_view_3.jpg} \\
    \end{tabular}} &
    \subfloat[SPSR]{%
    \begin{tabular}{c}
         \expandafter\includegraphics\expandafter[\trtCopts]{largescale_toronto/poisson_trim_view_3.jpg} \\
    \end{tabular}} &
    \subfloat[Ours]{%
    \begin{tabular}{c}
         \expandafter\includegraphics\expandafter[\trtCopts]{largescale_toronto/ours_view_3.jpg} \\
    \end{tabular}}
    \end{tabular}
    
    \caption{Evaluation of the methods on incomplete point clouds. The data is \textit{Toronto downtown}.}
    \label{fig:noise:missing}
\end{figure}

We evaluated the impact of additional noises (random noise and outliers) on CONet, SPSR, and our method on indoor scenes, as shown in Figure~\ref{fig:noise:additional}. The added noises may affect both the position of the points and the normals of the points. The middle column of Figure~\ref{fig:noise:additional} shows outliers introduced undesired objects to CONet, but it has much less impact on other methods and ours. However, noises can be problematic for SPSR since it relies on high-quality normal estimation, which can be easily affected by noises. CONet and our method have demonstrated a certain level of noises robustness in the reconstructed meshes. 
Incomplete data or missing data is a common problem in practice, such as those from laser scanning. Figure~\ref{fig:noise:missing} compares the results of CONet and ours in large-scale scenes, and we note that our approach is able to complete the missing facades with triangle faces in the incomplete region. Thanks to Delaunay triangulation and tetrahedron labeling, our method is able to close gaps and holes caused by missing data.

\subsection{Qualitative Evaluation\label{sec:quali}}
\begin{figure}
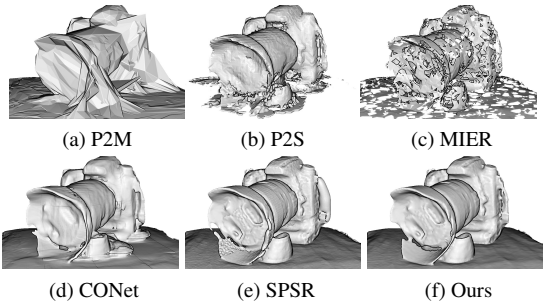

    \def\subimgwidth{0.28\linewidth}
    \edef\bullopts{trim=310 30 340 75,clip,width = \subimgwidth}
    \edef\triopts{trim=290 60 260 120,clip,width = \subimgwidth}
    \edef\grfopts{trim=370 0 510 15,clip,width = \subimgwidth}
    \edef\grfheadopts{trim=270 90 300 20,clip,width = \subimgwidth}
    \edef\tikiopts{trim=300 0 425 20,clip,width = \subimgwidth}
    \edef\DSLRViewOneopts{trim=180 100 280 50,clip,width = \subimgwidth}

    \centering
    \setlength{\tabcolsep}{2pt}

    \subfloat[P2M]{%
        \expandafter\includegraphics\expandafter[\DSLRViewOneopts]{object_dslr/point2mesh_view_1.jpg}} 
    \subfloat[P2S]{%
 \expandafter\includegraphics\expandafter[\DSLRViewOneopts]{object_dslr/p2s_view_1.jpg}} 
    \subfloat[MIER]{%
         \expandafter\includegraphics\expandafter[\DSLRViewOneopts]{object_dslr/mier_view_1.jpg} } \\ \vspace{-1em}
    \subfloat[CONet]{%
\expandafter\includegraphics\expandafter[\DSLRViewOneopts]{object_dslr/convonet_view_1.jpg} } 
    \subfloat[SPSR]{%
        \expandafter\includegraphics\expandafter[\DSLRViewOneopts]{object_dslr/poisson_view_1.jpg} } 
    \subfloat[Ours]{%
        \expandafter\includegraphics\expandafter[\DSLRViewOneopts]{object_dslr/ours_angle_view_1.jpg} }
\caption{Qualitative comparison on small object level datasets. The data is \textit{DSLR}.}
\label{fig:qual:smallobjsgrid}
\end{figure}

\begin{figure}
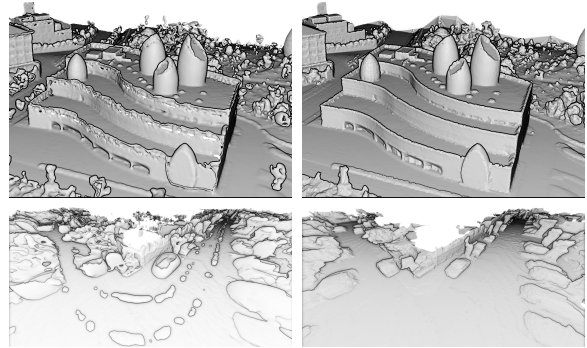

    \def\subimgwidth{0.45\linewidth}
    \edef\yparcopts{trim=150 50 150 50,clip,width = \subimgwidth}
    \edef\gsmopts{trim=100 0 100 0,clip,width = \subimgwidth}
    \edef\archwayopts{trim=100 80 150 220,clip,width = \subimgwidth}
    \edef\hotelopts{trim=100 20 300 10,clip,width = \subimgwidth}
    \edef\kittiopts{trim=100 20 300 10,clip,width = \subimgwidth}
    \edef\kittitwoopts{trim=5 20 5 160,clip,width = \subimgwidth}
    \edef\kittithreeopts{trim=5 20 5 20,clip,width = \subimgwidth}
    \edef\uthamopts{trim=50 10 50 100,clip,width = \subimgwidth}
    \edef\dragonparkopts{trim=250 200 450 110,clip,width = \subimgwidth}
    \edef\ecoparkopts{trim=280 00 150 100,clip,width = \subimgwidth}
    
    \centering
    \setlength{\tabcolsep}{1pt}
    \def\arraystretch{1}
    \begin{tabular}{cc}
    \subfloat[SPSR]{%
    \begin{tabular}{c}
        \expandafter\includegraphics\expandafter[\ecoparkopts]{ecopark/poisson_trim.jpg} \\
        \expandafter\includegraphics\expandafter[\kittithreeopts]{kitti_ablation/poisson_trim_view2.jpg}
    \end{tabular}} &
    \subfloat[Ours ($\lambda_{avw}=1$)]{%
    \begin{tabular}{c}
        \expandafter\includegraphics\expandafter[\ecoparkopts]{ecopark/ours_angle.jpg} \\
        \expandafter\includegraphics\expandafter[\kittithreeopts]{kitti_ablation/ours_view2.jpg} 
    \end{tabular}}
    \end{tabular}
    
    \caption{Qualitative comparison on the large-scale outdoor dataset. The data are \textit{Eco Park} (top) and \textit{Crossroad} (bottom). }
    \label{fig:qual:largescale}
\end{figure}

Our qualitative comparison is divided into three scales as well, shown in Figure~\ref{fig:qual:smallobjsgrid}, the first column of Figure~\ref{fig:noise:additional}, and Figure~\ref{fig:qual:largescale} respectively.
P2M~\cite{Hanocka2020p2m} and P2S~\cite{ErlerEtAl:Points2Surf:ECCV:2020} have poor reconstruction quality in terms of open surface (DSLR). Triangles generated by MIER~\cite{liu2020meshing} are close to the ground truth surface, but the completeness is poor. CONet usually produces a flat surface, which is a good feature for specific scenes (\ie indoors), but it is not good for curved surfaces. While SPSR~\cite{kazhdan2013screened} also produces reasonable reconstructions, it tends to close the resulting
meshes and present bubbles event after trimming with the threshold $max\_density/2$. A carefully chosen trimming parameter could reduce such artifacts. Moreover, instead of uniform sampling from the ground truth surface, we use real-world datasets derived from multi-view stereo~\cite{sensefly} or laser scan data~\cite{kitti2013} to demonstrate the generality of the proposed method. 
More results can be found in the supp. material.
\section{Conclusion}
We presented a novel surface reconstruction framework that combines the traditional graph-cut based method and learning-based method. In our method, the neural network only focuses on a very simple task - learning to predict the visibility of the point clouds given a view which thus present a much better generalization capability over existing learning-based approaches, while maintaining the efficiency of the traditional method. Specifically, we proposed a three-step network that explicitly employs depth completion for the visibility prediction of 3D points. Furthermore, we have improved graph-cut based formulation by proposing a novel adaptive visibility weighting term. Experiments show that our proposed method can be generalized to point clouds in different scene contexts, and can be extended to point clouds in large scenes. On datasets of varying scales, our framework presents better performance than the comparing classic reconstruction methods and state-of-the-art learning-based approaches. 

\boldparagraph{Acknowledgements}
Shuang Song and Rongjun Qin acknowledge the personnel support in part by Office of Naval Research (Award No.N000141712928) for this research effort. 
{\small
\bibliographystyle{ieee_fullname}
\bibliography{egbib}
}

\end{document}


\title{Vis2Mesh: Efficient Mesh Reconstruction from Unstructured Point Clouds of Large Scenes with Learned Virtual View Visibility\\
Supplementary Material}

\author{Shuang Song\\
The Ohio State University
\and
Zhaopeng Cui\thanks{Contribution initiated and performed at ETH Zürich.}\\
Zhejiang University
\and
Rongjun Qin\thanks{Corresponding author. Email: \small \href{mailto:qin.324@osu.edu}{qin.324@osu.edu}}\\
The Ohio State University
}

\maketitle
\ificcvfinal\thispagestyle{empty}\fi
\appendix



In this supplementary material, we first explain our graph weighting scheme with an example in Section~\ref{app:sec:graphweighting}. Then we present the details of our training dataset in  Section~\ref{app:sec:dataset}. In  Section~\ref{app:sec:frameworkandnetarch}, we describe the detailed network architecture. Next, we present the definition of the metrics for surface evaluation in Section~\ref{app:sec:metrics}. At last, we provide additional experimental results in Section~\ref{app:sec:extraexp}.

\section{Graph Weighting\label{app:sec:graphweighting}}
\begin{figure}
    \centering
    \includegraphics[width=\linewidth]{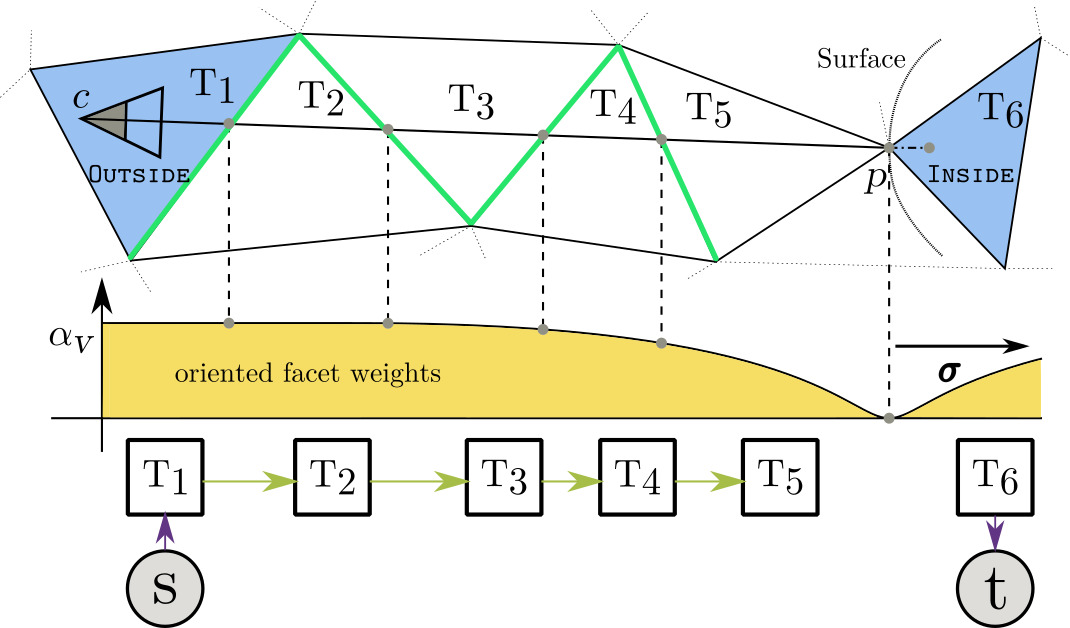}
    \caption{Visibility and graph construction~\cite{labatut2009robust}. From top to bottom, \textbf{Upper:} A line of sight from a reconstructed 3D point traverses a sequence of tetrahedra, the graph construction, and the assignment of weights to the tetrahedron and oriented facets. \textbf{Middle:} Soft visibility decay along the line of sight which is inversely proportional to the distance to the end point. \textbf{Lower:} Corresponding s-t graph and the cut solution.}
    \label{fig:softvisweighting}
\end{figure}

Corresponding to Section~\ref{msec:prelimiary} in the main paper, we present the tetrahedra graph weighting scheme of the basic method~\cite{labatut2009robust} by illustrating equations with their geometric meaning. As shown in Figure~\ref{fig:softvisweighting}, a line of sight traverse a series of tetrahedra (where $M=5$), from $T_1$ to $T_5$ and extended to $T_6$ ($T_{M+1}$ in the main paper) which is the tetrahedron right behind the end point $p$. The weighting process starts from the point of view ($c$) marking it as the source node (Equation~\ref{meq:origin_E_s} in the main paper). Along the line of sight, accumulate the value $\alpha_v$ to the connectivity of facets that line of sight passes through from the front (Equation~\ref{meq:origin_E_ij} in the main paper). As for $T_6$ behind the point, it is marked as a sink node with constant weight (Equation~\ref{meq:origin_E_t} in the main paper).

\section{Training Dataset\label{app:sec:dataset}}
\begin{figure}
    \centering
    \includegraphics[width=0.8\linewidth]{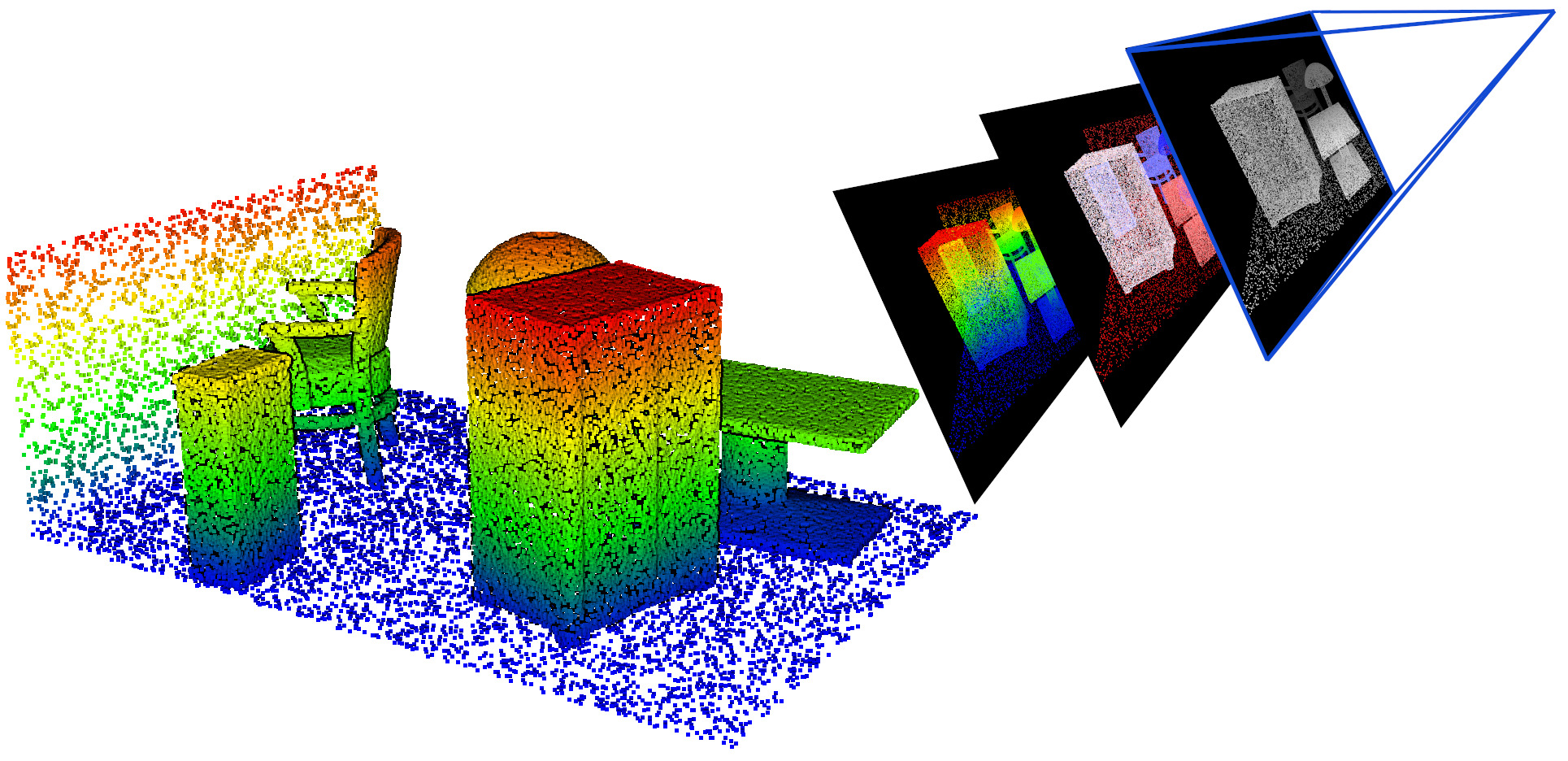}
    \caption{Record visibility information in an image by rendering: While points were rendered, the global index of the points been recorded simultaneously.}
    \label{fig:method_rendering}
\end{figure}

Our training dataset is built based on textured mesh models of public Multi-View Stereo (MVS) reconstruction dataset BlendedMVS~\cite{yao2020blendedmvs}. We select 5 textured models as ground truth surfaces for training and 2 models for testing. 
For each model, we uniformly sample points from the surface. The number of sampled points depends on the actual size and complexity of the structure, ranging from 500K to 10M points. We prefer sparse sampling since the more occluded points have been projected to the virtual views, the more contrastive samples we will have. We collect datasets by sampling virtual views (Section~\ref{msec:viewgeneratos} in the main paper) and rendering them with our renderer. 
Our renderer is implemented with OpenGL, which not only provides regular color images and depth images but also records the original index of each projection point, as shown in Figure~\ref{fig:method_rendering}.

\begin{figure}
    \def\subimgwidth{0.45\linewidth}
    \centering
    \setlength{\tabcolsep}{2pt}
    \def\arraystretch{0}
    \begin{tabular}{cc}
    \subfloat[Ground truth surface]{%
        \includegraphics[width=\subimgwidth]{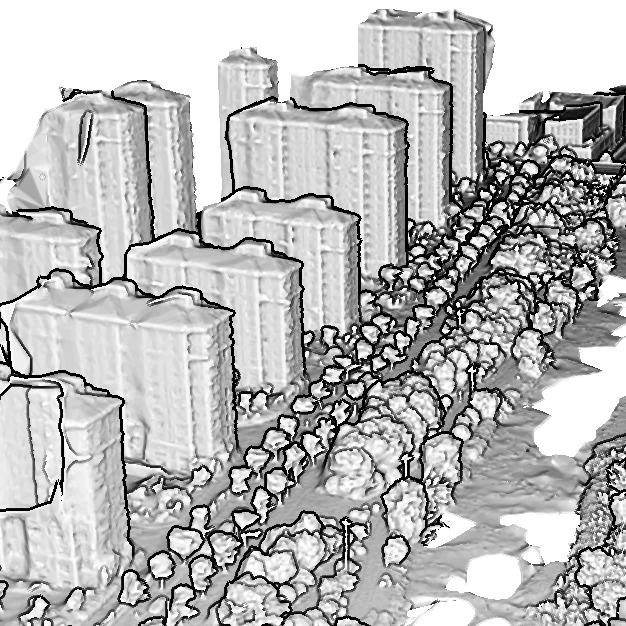}}&
    \subfloat[Sampled point cloud]{%
        \includegraphics[width=\subimgwidth]{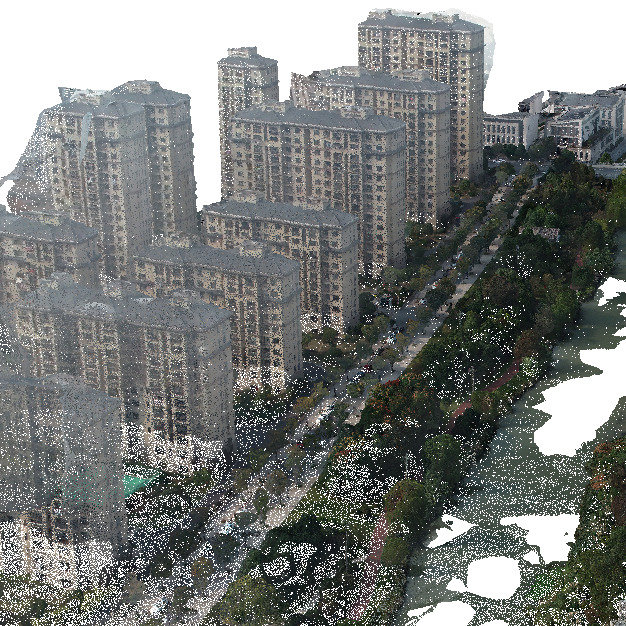}}\\
    \subfloat[Surface-rendering depth map]{%
        \includegraphics[width=\subimgwidth]{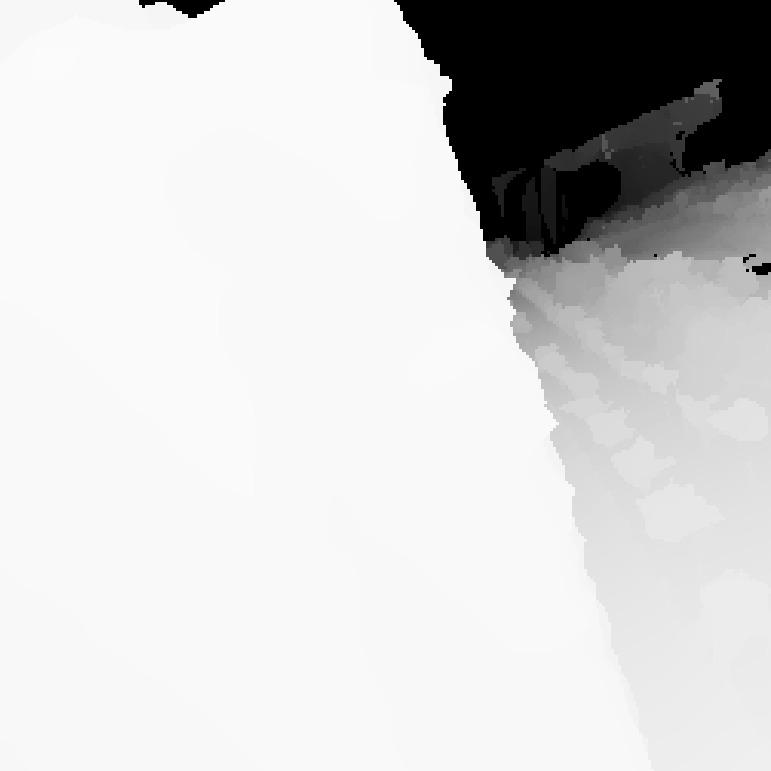}} &
    \subfloat[Point-rendering color image]{%
      \includegraphics[width=\subimgwidth]{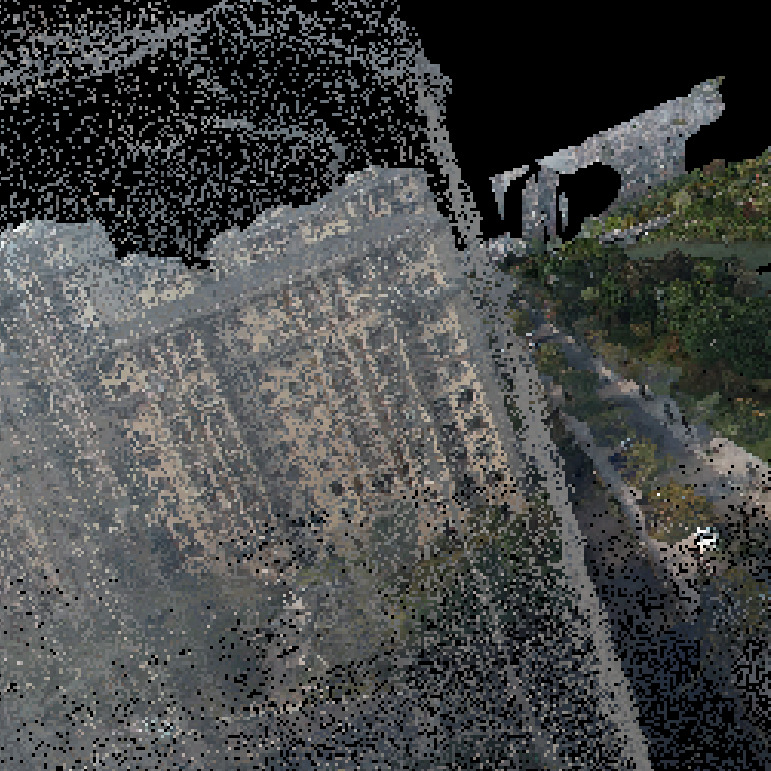}} \\
    \subfloat[Point-rendering ID map]{%
        \includegraphics[width=\subimgwidth]{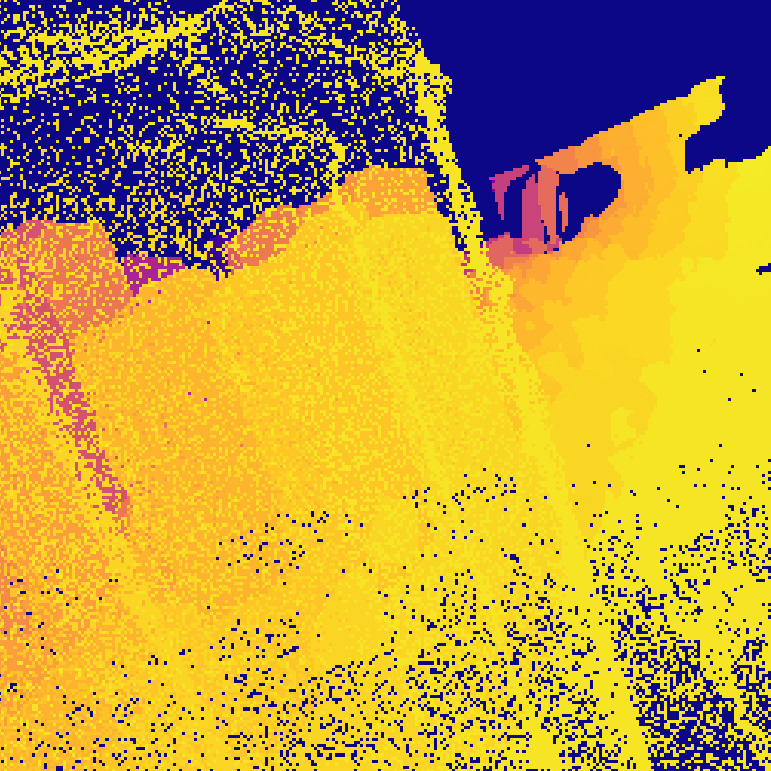}} &
    \subfloat[Point-rendering depth map]{%
        \includegraphics[width=\subimgwidth]{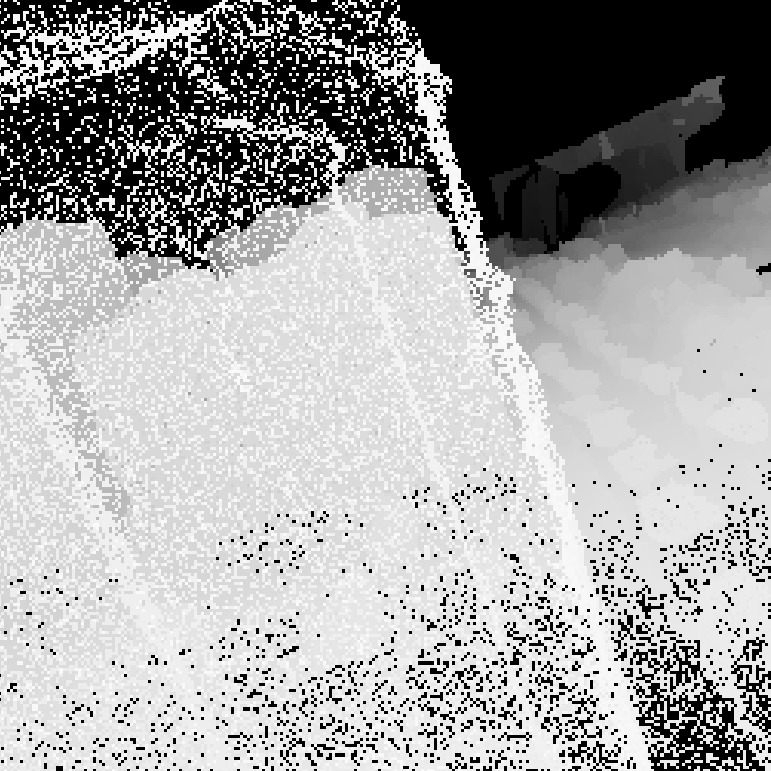}} \\
    \subfloat[Ground truth visibility]{%
        \includegraphics[width=\subimgwidth]{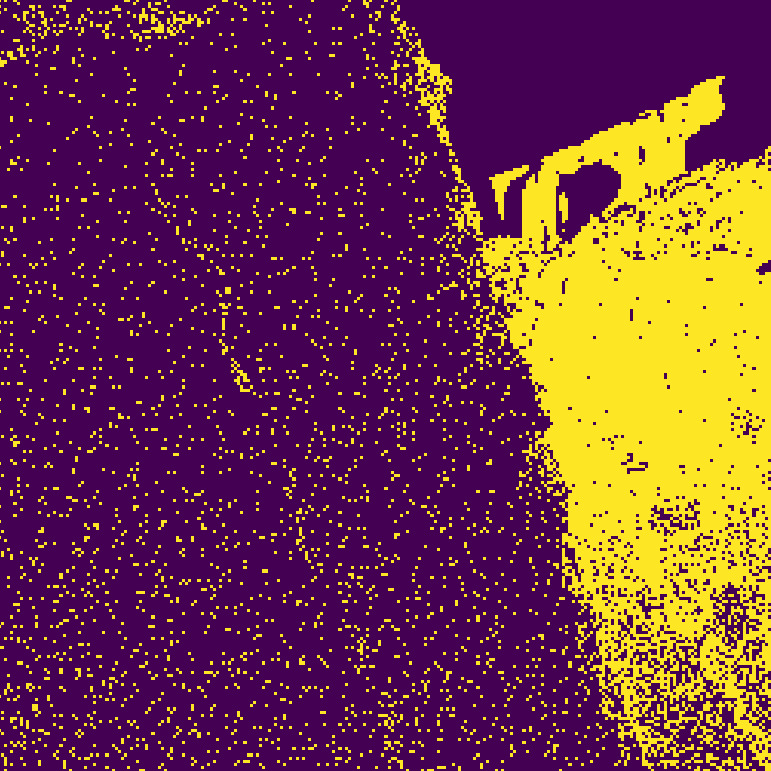}}&
    \subfloat[Ground truth cleaned depth map]{%
        \includegraphics[width=\subimgwidth]{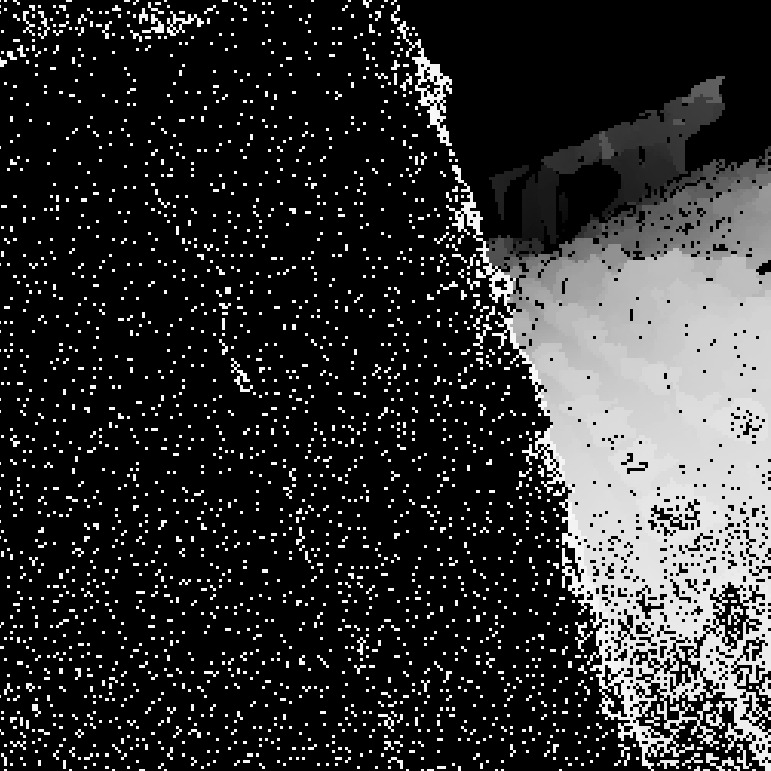}} \\
    \end{tabular}
  
  \caption{The content of our dataset.}
  \label{fig:ourdatasetinspection} 
\end{figure}

Figure~\ref{fig:ourdatasetinspection} shows an example out of 1414 virtual views. The ground truth visibility is generated by comparing the point-rendering depth map and the surface-rendering depth map. If the difference of depth is less than $\epsilon=0.05$, it will be marked as visible, otherwise, it will be marked as occluded.

\section{Architecture\label{app:sec:frameworkandnetarch}}
Table~\ref{tab:modulelist} provides a detailed description of the input and output tensor sizes of each module in our workflow. And Table~\ref{tab:detailunet} presents the detailed architecture of the network, including the size of each buffer.

\begin{table*}
    \centering
    \begin{tabular}{|c|c|l|c|}
    \hline
         \textbf{Module} & \textbf{IO} & \textbf{Description} & \textbf{Data Dimension}\\\hline\hline
         
         \multirow{3}{*}{Virtual View Sampler} & Input & Given Point Cloud & $n \times 3$ \\
         & Input & The flag indicate the pattern of view generator & 1 \\
         & Output & 6 DoF poses & $m \times 6\times 1$ \\\hline
         
         \multirow{5}{*}{Renderer} &  Input & Given Point Cloud & $n \times 3$ \\
         & Input & 6 DoF pose & $6\times 1$ \\
         & Output & Rendered sparse color image & $H \times W \times 3$ \\
         & Output & Rendered sparse depth map & $H\times W\times 1$ \\
         & Output & Rendered sparse point ID map & $H\times W\times 1$ \\\hline
         
         \multirow{2}{*}{CoarseVisNet} & Input & Normalized and binary mask attached sparse depth map & $H\times W\times 2$ \\
         & Output & Predicted mask of visible pixels & $H\times W\times 1$ \\\hline
         
         \multirow{3}{*}{Multiplication} & Input & Rendered sparse depth map & $H\times W \times 1$ \\
         & Input & Predicted mask of visible pixels & $H\times W\times 1$ \\
         & Output & Cleaned sparse depth map & $H\times W\times 1$  \\\hline
         
         \multirow{2}{*}{DepthCompNet} & Input & Normalized and binary mask attached cleaned sparse depth map & $H\times W\times 2$ \\
         & Output & Completed dense depth map & $H\times W\times 1$ \\\hline
         
         \multirow{3}{*}{Concatenation} & Input & Normalized and binary mask attached sparse depth map & $H\times W\times 2$  \\
         & Input & Normalized and binary mask attached cleaned sparse depth map & $H\times W\times 1$  \\
         & Output & Concatenated raw depth and completed depth map & $H\times W\times 3$  \\ \hline
         
         \multirow{2}{*}{FineVisNet} & Input & Concatenated raw depth and completed depth map & $H\times W\times 3$ \\
         & Output & Predicted mask of visible pixels & $H\times W\times 1$ \\\hline
         
         Graph-cut & Input & Given Point Cloud & $n \times 3$ \\
         based & Input & All rendered sparse point ID map & $m\times H\times W \times 1$ \\
         Surface & Input & Predicted mask of visible pixels & $m\times H\times W\times 1$ \\
         Reconstruction & Output & Reconstructed triangle surface & Irregular \\
         
         \hline
    \end{tabular}
    \vspace{1em}
    \caption{Modules overview. We detail the input and output of all modules that appeared in our workflow. $n$ is the number of input points, $m$ is the number of generated virtual views. $H$ and $W$ are the customized values which are set to $256 \times 256$ in our experiments.}
    \vspace{1em}
    \label{tab:modulelist}
\end{table*}

\begin{table*}[htp]
    \centering
    \begin{tabular}{|c|c|c|c|}
    \hline
         \textbf{Part} & \textbf{Layer} & \textbf{Parameters} & \textbf{Output Dimension}  \\\hline\hline
         \multirow{5}{*}{Encoder}& Double PConv1BnReLU & 3x3,64,64 & $H\times W\times 64$ \\
         & MaxPool + Double PConv2BnReLU & 2,3x3,128,128 & $H/2\times W/2\times 128$ \\
         & MaxPool + Double PConv3BnReLU & 2,3x3,256,256 & $H/4\times W/4\times 256$ \\
         & MaxPool + Double PConv4BnReLU & 2,3x3,512,512 & $H/8\times W/8\times 512$ \\
         & MaxPool + Double PConv5BnReLU & 2,3x3,512,512 &
         $H/16\times W/16\times 512$ \\
         \hline
         \multirow{5}{*}{Decoder} & Bilinear Upsample1 & 2 & $H/8\times W/8\times 512$ \\
         & Concat1 & cat(PConv4,Unsample1) & $H/8\times W/8\times 1024$ \\
         & Double PConv6BnReLU & 3x3,256,256 & $H/8\times W/8\times 256$ \\
         & Bilinear Upsample2 & 2 & $H/4\times W/4\times 256$ \\
         & Concat2 & cat(PConv3,Unsample2) & $H/4\times W/4\times 512$ \\
         & Double PConv7BnReLU & 3x3,128,128 & $H/4\times W/4\times 128$ \\
         & Bilinear Upsample3 & 2 & $H/2\times W/2\times 128$ \\
         & Concat3 & cat(PConv2,Unsample3) & $H/2\times W/2\times 256$ \\
         & Double PConv8BnReLU & 3x3,64,64 & $H/2\times W/2\times 64$ \\
         & Bilinear Upsample4 & 2 & $H\times W\times 64$ \\
         & Concat4 & cat(PConv1,Unsample4) & $H\times W\times 128$ \\
         & Double PConv9BnReLU & 3x3,64,64 & $H\times W\times 64$ \\
         & Sigmoid & - & $H\times W\times 1$ \\
         \hline
    \end{tabular}
    \vspace{1em}
    \caption{Detailed encoder-decoder network architecture used for our \textbf{CoarseVisNet}, \textbf{DepthCompNet}, and \textbf{FineVisNet}. \textbf{PConv}: partial convolution layer. \textbf{Double PConvBnReLU}: PConv+BatchNorm+ReLU+PConv+BatchNorm+ReLU. Parameter of \textbf{Double PConvBnReLU}: kernel size, number of filters for the first PConv, and the second PConv.}
    \label{tab:detailunet}
\end{table*}

\section{Surface Evaluation Metrics\label{app:sec:metrics}}

The quantitative comparison of two mesh surfaces is a complicated problem. Alternatively, we employ highly oversampled points of triangle surfaces to approximate mesh to mesh metrics. In our experiment, we control the number of points by the density of sampling. Furthermore, to eliminate the randomness introduced by the sampling method, the ground truth mesh is sampled twice to determine the scale factors for the following metrics.

\boldparagraph{Chamfer distance (CD)~\cite{barrow1977parametric}}. For each point, we find the nearest neighbor in the other set and sums the distances up. To ensure Chamfer distance as symmetric, we sum distances of $\mathcal{P}$ to $\mathcal{Q}$ and $\mathcal{Q}$ to $\mathcal{P}$. $K$ is a scale factor to leverage between different datasets: $K=\frac{1}{CD(\mathcal{S}_1,\mathcal{S}_2)}$.

\begin{equation}
    \small \text{CD}(\mathcal{P}, \mathcal{Q}) = \frac{K}{\lvert \mathcal{P} \rvert}\sum_{p \in \mathcal{P}} \min_{q \in \mathcal{Q}} \lVert p-q \rVert_2 + \frac{K}{\lvert \mathcal{Q} \rvert}\sum_{q \in \mathcal{Q}} \min_{p \in \mathcal{P}} \lVert p-q \rVert_2 .
    \label{eq:chamferdistance}
\end{equation}

\boldparagraph{F-score~\cite{tanksandtemples}}. As shown in Equation~\ref{eq:fscorecombo}, the F-score is defined as the harmonic mean between precision and recall. $\mathcal{P} \subseteq \mathbb{
R}^3$ is the point set sampled from the generated surface while $\mathcal{Q} \subseteq \mathbb{
R}^3$ is the point set sampled from the ground truth surface. Threshold $T$ is determined by computing maximum distance between two point sets ($\mathcal{S}_{1,2} \subseteq \mathbb{
R}^3$) randomly sampled from the same ground truth surface: $T=\max_{s_1\in \mathcal{S}_1}(\min_{s_2\in \mathcal{S}_2} \lVert s_1 - s_2 \rVert_2)$.

{\small
\begin{subequations}
\begin{align}
    \text{Precision}(\mathcal{P},\mathcal{Q}) &= \frac{ \sum_{q \in \mathcal{Q}} \delta[\min_{p\in \mathcal{P}}\lVert p - q \rVert_2 < T] }{\lvert \mathcal{Q} \rvert} ,\label{eq:precision} \\
    \text{Recall}(\mathcal{P},\mathcal{Q}) &= \frac{ \sum_{p \in \mathcal{P}} \delta[ \min_{q\in \mathcal{Q}} \lVert p-q \rVert_2 <T ] }{\lvert \mathcal{P} \rvert}\label{eq:recall} ,\\
    \text{F-score}(\mathcal{P}, \mathcal{Q}) &= 2 \cdot \frac{\text{Precision} \cdot \text{Recall}}{\text{Precision} + \text{Recall}} .\label{eq:fscore}
\end{align}
\label{eq:fscorecombo}
\end{subequations}
}

\section{Additional Experimental Results\label{app:sec:extraexp}}
In this section, we provide additional experimental results for better evaluation of our method.
Table~\ref{tab:datasetsheet} shows the data volume, source, and corresponding view generator for all the datasets we used in our experiments. We can see that different scales of data are evaluated in our experiment. 

\begin{table}
    \centering
    \resizebox{1\linewidth}{!}{
    \begin{tabular}{|l|l|l|l|l|}
    \hline
    \textbf{Name} & \textbf{\# Points} & \textbf{Scale} & \textbf{Source} & \textbf{Generator}  \\\hline\hline
    Triceratops & 25K & Object & P2M & Spherical \\
    Tiki & 5K & Object & P2M & Spherical \\
    Giraffe & 25K & Object & COSEG & Spherical \\
    Bull & 25K & Object & COSEG & Spherical \\
    DSLR & 25K & Object & BlendedMVS & Spherical \\
    Birdcage & 25K & Object & Thingi10k & Spherical \\
    Room0 & 100K & Indoor & CONet & User \\
    Room1 & 100K & Indoor & CONet & User \\    
    MPT:SingleFloor & 400K & Indoor & Matterport & User \\
    MPT:MultiFloor & 500K & Indoor & Matterport & User \\
    Toronto Downtown \#1 & 177K & Outdoor & ISPRS & Nadir \\
    Toronto Downtown \#2 & 300K & Outdoor & ISPRS & Nadir \\
    Columbus City & 925K & Outdoor & OGRIP & Nadir \\
    Church & 500K & Outdoor & BlendedMVS & Oblique \\
    Archway & 500K & Outdoor & BlendedMVS & Oblique \\
    Dragon Park & 500K & Outdoor & BlendedMVS & Oblique \\
    Dragon Park (Teaser) & 5M & Outdoor & BlendedMVS & Oblique \\
    Eco Park & 500K & Outdoor & BlendedMVS & Oblique \\
    Pedestrian street & 500K & Outdoor & BlendedMVS & Oblique \\
    YParc & 2M & Outdoor & senseFly & Nadir \\
    UThammasat & 5M & Outdoor & senseFly & Nadir \\
    Hotel & 5M & Outdoor & senseFly & Oblique \\
    GSM Tower & 3.3M & Outdoor & senseFly & User \\
    Street View & 5M & Outdoor & MAI & User \\
    Crossroad & 2.6M & Outdoor & KITTI & User \\
    \hline
    \end{tabular}
    }
    \vspace{0.5em}
    \caption{Details of data of our experiments, including their source, volume of the data, as well as virtual view generator.}
    \label{tab:datasetsheet}
\end{table}

\subsection{Extra Evaluation of Visibility Classifiers\label{app:sec:quantitative:classifiers}}
In this section, we provide test scores on two more datasets in Table~\ref{app:tab:network_quantitative_mvs} and Table~\ref{app:tab:network_quantitative_simlidar} to evaluate our VDVNet with  datasets of different modalities, which are multi-view stereo dataset and LiDAR scanning dataset.

\begin{table}[htp]
    \centering
    \resizebox{\linewidth}{!}{
        \begin{tabular}{|c|c|c|c|c|c|}
        \hline
        \textbf{Visibility Estimator} & \%P & \%R & \%F1 & \%AUC \\
        \hline\hline
        HPR  & 90.12&	79.58	&86.2&	75.81\\ \hline
        UNet  & 89.54	&84.21&	86.6&	75.89 \\
        UNet + PConv  & \textbf{90.23}&	84.7&	87.18&	78.09\\ 
        \textsc{VisibNet}  & 89.13&	86.74&	87.74&	75.8 \\ \hline
        Ours w/o DepthComp or PConv  & 89.49&	86.43&	87.75&	77.37\\
        Ours w/o DepthComp  & 89.65	&90.25&	89.83&	78.73\\
        Ours w/o PConv  & 89.6&	89.28&	89.33&	78.17\\
        Ours  & 90.09&	\textbf{93.3}&	\textbf{91.52}&	\textbf{81.59} \\
        \hline
        \end{tabular}
        }
    \caption{Quantitative analysis of methods on binary visibility classification task on MVS dataset.}
    \label{app:tab:network_quantitative_mvs}
\end{table}

\begin{table}[htp]
    \centering
    \resizebox{\linewidth}{!}{
        \begin{tabular}{|c|c|c|c|c|c|}
        \hline
        \textbf{Visibility Estimator} & \%P & \%R & \%F1 & \%AUC \\
        \hline\hline
        HPR  & 80.6	&85.48&	82.14&	81.29\\ \hline
        UNet  & 78.71&	75.58&	76.6&	79.84 \\
        UNet + PConv  & 77.98&	82.49&	79.67&	81.17\\ 
        \textsc{VisibNet}  & 76.98&	85.59&	80.52&	80.76\\ \hline
        Ours w/o DepthComp or PConv  & 77.17&	85.18&	80.26&	79.11\\
        Ours w/o DepthComp  & 78.00 &	87.23&	81.81&	81.69\\
        Ours w/o PConv  & 79.29&	91.29&	84.31&	83.57\\
        Ours  & \textbf{80.4}&	\textbf{93.71}&	\textbf{85.27}&	\textbf{83.96} \\
        \hline
        \end{tabular}
        }
    \caption{Quantitative analysis of methods on binary visibility classification task on Simulated LiDAR dataset.}
    \label{app:tab:network_quantitative_simlidar}
\end{table}
\subsection{Performance on Extreme Sparse Data\label{app:sec:extrasparse}}
\begin{figure}[htp]
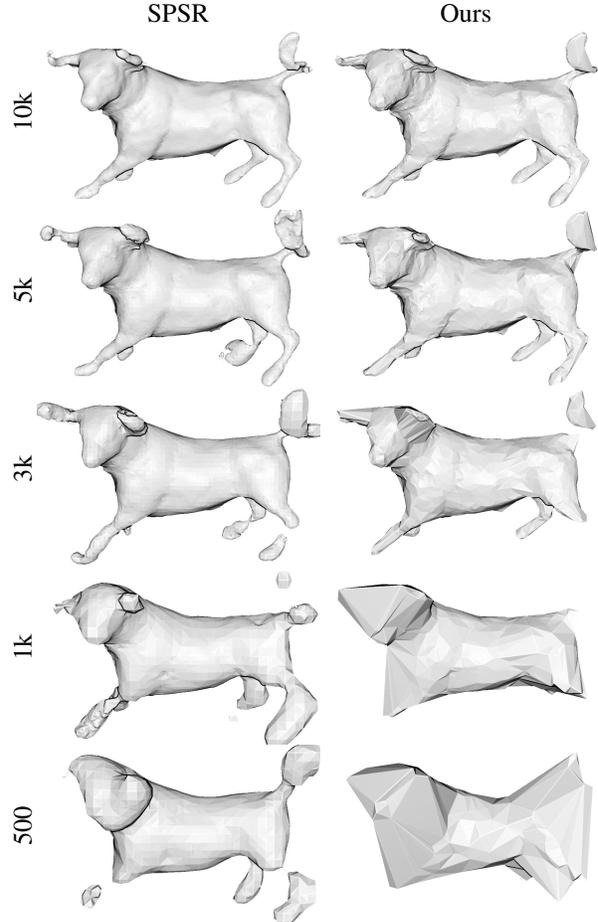

    \def\subimgwidth{0.45\linewidth}
    \edef\bullopts{trim=140 70 220 70,clip,width = \subimgwidth}
    
    \centering
    \setlength\tabcolsep{1pt}
    \def\arraystretch{1}
    \begin{tabular}{ccc}
        & SPSR & Ours \\
        \rotatebox[origin=c]{90}{10k} &
        \raisebox{-0.5\height}{\expandafter\includegraphics\expandafter[\bullopts]{extremesparse/p10k.jpg}} &
        \raisebox{-0.5\height}{\expandafter\includegraphics\expandafter[\bullopts]{extremesparse/o10k.jpg}} \\
        \rotatebox[origin=c]{90}{5k} &
        \raisebox{-0.5\height}{\expandafter\includegraphics\expandafter[\bullopts]{extremesparse/p5k.jpg}} &
        \raisebox{-0.5\height}{\expandafter\includegraphics\expandafter[\bullopts]{extremesparse/o5k.jpg}} \\
        \rotatebox[origin=c]{90}{3k} &
        \raisebox{-0.5\height}{\expandafter\includegraphics\expandafter[\bullopts]{extremesparse/p3k.jpg}} &
        \raisebox{-0.5\height}{\expandafter\includegraphics\expandafter[\bullopts]{extremesparse/o3k.jpg}} \\
        \rotatebox[origin=c]{90}{1k} &
        \raisebox{-0.5\height}{\expandafter\includegraphics\expandafter[\bullopts]{extremesparse/p1k.jpg}} &
        \raisebox{-0.5\height}{\expandafter\includegraphics\expandafter[\bullopts]{extremesparse/o1k.jpg}} \\
        \rotatebox[origin=c]{90}{500} &
        \raisebox{-0.5\height}{\expandafter\includegraphics\expandafter[\bullopts]{extremesparse/p500.jpg}} &
        \raisebox{-0.5\height}{\expandafter\includegraphics\expandafter[\bullopts]{extremesparse/o500.jpg}} \\
    \end{tabular}
    \caption{Performance on a series of point clouds with different densities.}
    \label{fig:extremesparse}
\end{figure}

As shown in Figure~\ref{fig:extremesparse}, We used a series of downsampled bull datasets to evaluate the performance of our method on very low-density points. The reconstruction quality of our method drops obviously as long as the number of points decreasing. It is because the initial mesh created by Delaunay is a bottom-up approach, and the graph-cut only applies a smooth constraint. The core idea of the proposed method is to recover details in the mesh reconstruction process based on correct visibility recovery, where these original point measurements are available. Therefore having sufficient points will show this advantage.

\subsection{Efficiency Statistics\label{app:sec:efficiency}}
We report the processing time for each component of our method in Table~\ref{tab:effcomp} and compare it with the SPSR method.
Efficiency performance is evaluated on the Laptop with Intel-4700MQ, 16 GB RAM, and Nvidia 970M (6GB GRAM), which is a challenge on such a resource-limited platform for reconstruction approaches. 
The processing time of our method increases with the number of points and virtual views, and the SPSR is related to the depth of the octree. In our experiment, we decide the number of views by the complexity of the scene, and we set the depth to 10 levels for SPSR. In addition, our VDVNet takes 993MB GPU RAM for visibility estimation (image size: 256x256).

\begin{table*}
    \centering
    \resizebox{0.8\linewidth}{!}{
    \begin{tabular}{|c|c|c||c|c|c||c|c||c|c|}
    \hline
    \multirow{2}{*}{\textbf{Data}} & \textbf{\# of}  & \textbf{\# of virtual} & \textbf{Visibility}  & \textbf{Graph-based} & \textbf{Overall} & \textbf{Normal} &\textbf{SPSR} & \textbf{Peak RAM} & \textbf{Peak RAM} \\
    & \textbf{points} & \textbf{views} & \textbf{estimation} & \textbf{reconstruction} & \textbf{(ours)} & \textbf{estimation} & \textbf{} & \textbf{(ours)} & \textbf{(SPSR)}\\
    
    \hline\hline
    Object (Quantitative) & 25K & $\sim$30 & $\sim$3s & $\sim$1s & $\sim$4s & $\sim$1s & $\sim$3s & $\sim$0.1GB & $\sim$0.2GB \\
    Indoor (Quantitative) & 100K & $\sim$50 & $\sim$8s & $\sim$4s & $\sim$12s & $\sim$3s & $\sim$6s & $\sim$0.2GB & $\sim$0.3GB \\
    Outdoor (Quantitative) & 500K & $\sim$100 & $\sim$14s & $\sim$13s & $\sim$27s & $\sim$4s & $\sim$22s & $\sim$0.8GB & $\sim$1.8GB \\
    YParc & 2M & 203 & 24s & 152s & 179s & 15s & 192s & 3.4GB & 5.2GB\\
    UThammasat & 5M & 300 & 30s & 278s & 311s &38s & 204s & 4.3GB& 5.9GB\\
    Hotel & 5M & 340 & 38s & 229s & 272s & 40s & 208s & 4.1GB & 5.9GB \\
    GSM Tower & 3.3M & 87 & 10s & 92s & 103s & 23s & 195s & 4.2GB & 5.1GB\\
    Crossroad & 2.6M & 114 & 15s & 71s & 86s & 12s & 105s & 3.2GB & 4.6GB\\
    \hline
    \end{tabular}}
    \caption{Computational efficiency comparison with SPSR. To note that our overall processing time including image rendering, visibility estimation, and graph-cut based reconstruction.}
    \label{tab:effcomp}
\end{table*}

\subsection{Qualitative Evaluation of Visibility Classifiers\label{app:sec:qual:classifiers}}
\begin{figure}[ht]
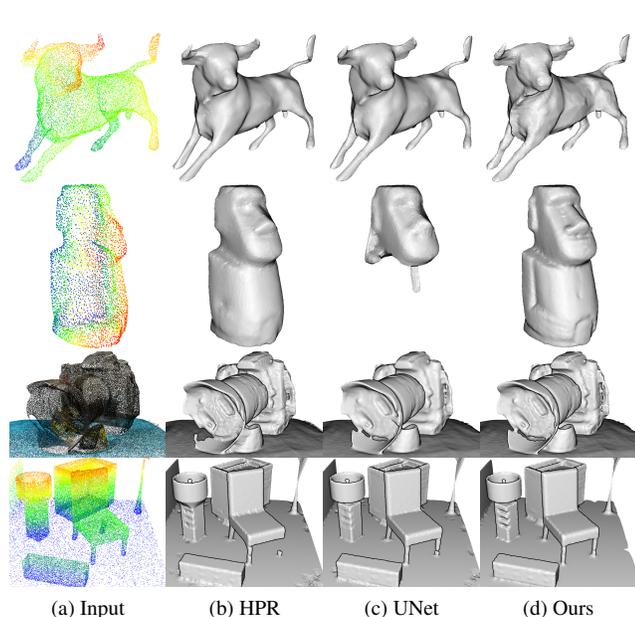

    \def\subimgwidth{0.25\linewidth}
    \edef\bullopts{trim=310 30 340 75,clip,width = \subimgwidth}
    \edef\triopts{trim=290 60 260 120,clip,width = \subimgwidth}
    \edef\grfopts{trim=370 0 510 15,clip,width = \subimgwidth}
    \edef\grfheadopts{trim=320 80 350 20,clip,width = \subimgwidth}
    \edef\tikiopts{trim=250 0 385 20,clip,width = \subimgwidth}
    \edef\roomZeroViewOneopts{trim=350 0 230 165,clip,width = \subimgwidth}
    \edef\roomOneViewTwoopts{trim=300 30 230 50,clip,width = \subimgwidth}
    \edef\DSLRViewOneopts{trim=200 100 300 50,clip,width = \subimgwidth}
    \centering
    \setlength{\tabcolsep}{0pt}
    \def\arraystretch{0}
    \begin{tabular}{cccc} 
    \subfloat[Input]{%
    \begin{tabular}{c}
         \expandafter\includegraphics\expandafter[\bullopts]{object_bull/pointcloud_ramp_view_1.jpg} \\
         \expandafter\includegraphics\expandafter[\tikiopts]{object_tiki/pointcloud_ramp_view_1.jpg} \\
         \expandafter\includegraphics\expandafter[\DSLRViewOneopts]{object_dslr/pointcloud_view_1.jpg} \\
         \expandafter\includegraphics\expandafter[\roomOneViewTwoopts]{indoor_room_1_raw/pointcloud_ramp_view_2.jpg}
    \end{tabular}} &
    \subfloat[HPR]{%
    \begin{tabular}{c}
         \expandafter\includegraphics\expandafter[\bullopts]{object_bull/hpr_view_1.jpg} \\ 
         \expandafter\includegraphics\expandafter[\tikiopts]{object_tiki/hpr_view_1.jpg} \\
         \expandafter\includegraphics\expandafter[\DSLRViewOneopts]{object_dslr/hpr_view_1.jpg} \\
         \expandafter\includegraphics\expandafter[\roomOneViewTwoopts]{indoor_room_1_raw/hpr_view_2.jpg}
    \end{tabular}} &
    \subfloat[UNet]{%
    \begin{tabular}{c}
         \expandafter\includegraphics\expandafter[\bullopts]{object_bull/unet_view_1.jpg} \\
         \expandafter\includegraphics\expandafter[\tikiopts]{object_tiki/unet_view_1.jpg} \\
         \expandafter\includegraphics\expandafter[\DSLRViewOneopts]{object_dslr/unet_view_1.jpg} \\
         \expandafter\includegraphics\expandafter[\roomOneViewTwoopts]{indoor_room_1_raw/unet_view_2.jpg}
    \end{tabular}} &
    \subfloat[Ours]{%
    \begin{tabular}{c}
        \expandafter\includegraphics\expandafter[\bullopts]{object_bull/ours_view_1.jpg} \\
        \expandafter\includegraphics\expandafter[\tikiopts]{object_tiki/ours_view_1.jpg} \\
        \expandafter\includegraphics\expandafter[\DSLRViewOneopts]{object_dslr/ours_angle_view_1.jpg} \\
        \expandafter\includegraphics\expandafter[\roomOneViewTwoopts]{indoor_room_1_raw/ours_angle_view_2.jpg}
    \end{tabular}}
    \end{tabular}
    
    \caption{Qualitative comparison of visibility estimators. From top to bottom, the data is \textit{Bull}, \textit{Tiki}, \textit{DSLR}, and \textit{Room1}. }
    \label{fig:recon:estimators}
\end{figure}

Figure~\ref{fig:recon:estimators} shows the qualitative evaluation of different visibility estimators and visually shows the result in Table~\ref{mtab:network_quantitative} in our main paper. The improvement in classification accuracy is reflected in the sharpness of the reconstructed surface details.

\subsection{Extra Robust Evaluation\label{app:sec:extra:noise}}
\begin{figure}
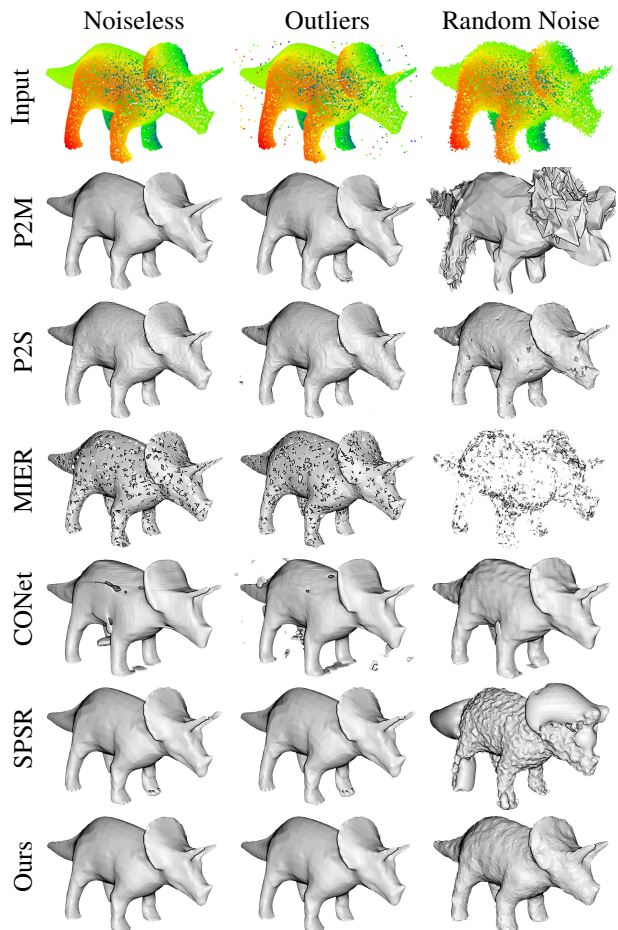

    \def\subimgwidth{0.3\linewidth}
    \edef\trimopt{trim=290 60 260 120,clip,clip,width = \subimgwidth}
    
    \centering
    \setlength\tabcolsep{1pt}
    \def\arraystretch{1}
    \begin{tabular}{cccc}
        & Noiseless & Outliers & Random Noise \\
        \rotatebox[origin=c]{90}{Input} &
        \raisebox{-0.5\height}{\expandafter\includegraphics\expandafter[\trimopt]{object_triceratops_raw/pointcloud_ssao_view_1.jpg}} &
        \raisebox{-0.5\height}{\expandafter\includegraphics\expandafter[\trimopt]{object_triceratops_outlier/pointcloud_ssao_view_1.jpg}} &
        \raisebox{-0.5\height}{\expandafter\includegraphics\expandafter[\trimopt]{object_triceratops_noise/pointcloud_ssao_view_1.jpg}} \\
        \rotatebox[origin=c]{90}{P2M} &
        \raisebox{-0.5\height}{\expandafter\includegraphics\expandafter[\trimopt]{object_triceratops_raw/point2mesh_view_1.jpg}} &
        \raisebox{-0.5\height}{\expandafter\includegraphics\expandafter[\trimopt]{object_triceratops_outlier/point2mesh_view_1.jpg}} &
        \raisebox{-0.5\height}{\expandafter\includegraphics\expandafter[\trimopt]{object_triceratops_noise/point2mesh_view_1.jpg}} \\
        \rotatebox[origin=c]{90}{P2S} &
        \raisebox{-0.5\height}{\expandafter\includegraphics\expandafter[\trimopt]{object_triceratops_raw/p2s_view_1.jpg}} &
        \raisebox{-0.5\height}{\expandafter\includegraphics\expandafter[\trimopt]{object_triceratops_outlier/p2s_view_1.jpg}} &
        \raisebox{-0.5\height}{\expandafter\includegraphics\expandafter[\trimopt]{object_triceratops_noise/p2s_view_1.jpg}} \\
        \rotatebox[origin=c]{90}{MIER} &
        \raisebox{-0.5\height}{\expandafter\includegraphics\expandafter[\trimopt]{object_triceratops_raw/mier_view_1.jpg}} &
        \raisebox{-0.5\height}{\expandafter\includegraphics\expandafter[\trimopt]{object_triceratops_outlier/mier_view_1.jpg}} &
        \raisebox{-0.5\height}{\expandafter\includegraphics\expandafter[\trimopt]{object_triceratops_noise/mier_view_1.jpg}} \\
        \rotatebox[origin=c]{90}{CONet} &
        \raisebox{-0.5\height}{\expandafter\includegraphics\expandafter[\trimopt]{object_triceratops_raw/convonet_view_1.jpg}} &
        \raisebox{-0.5\height}{\expandafter\includegraphics\expandafter[\trimopt]{object_triceratops_outlier/convonet_view_1.jpg}} &
        \raisebox{-0.5\height}{\expandafter\includegraphics\expandafter[\trimopt]{object_triceratops_noise/convonet_view_1.jpg}} \\
        \rotatebox[origin=c]{90}{SPSR} &
        \raisebox{-0.5\height}{\expandafter\includegraphics\expandafter[\trimopt]{object_triceratops_raw/poisson_view_1.jpg}} &
        \raisebox{-0.5\height}{\expandafter\includegraphics\expandafter[\trimopt]{object_triceratops_outlier/poisson_view_1.jpg}} &
        \raisebox{-0.5\height}{\expandafter\includegraphics\expandafter[\trimopt]{object_triceratops_noise/poisson_view_1.jpg}} \\
        \rotatebox[origin=c]{90}{Ours} &
        \raisebox{-0.5\height}{\expandafter\includegraphics\expandafter[\trimopt]{object_triceratops_raw/ours_view_1.jpg}} &
        \raisebox{-0.5\height}{\expandafter\includegraphics\expandafter[\trimopt]{object_triceratops_outlier/ours_view_1.jpg}} &
        \raisebox{-0.5\height}{\expandafter\includegraphics\expandafter[\trimopt]{object_triceratops_noise/ours_view_1.jpg}} 
    \end{tabular}
    \caption{Evaluation of the methods subject to additional noises on the point clouds. The data is \textit{Triceratops}.}
    \label{fig:robustness}
\end{figure}

In Figure~\ref{fig:robustness} we present extra results for robustness evaluation of noises (Figure~\ref{mfig:noise:additional} in the main paper). 

\subsection{Extra Qualitative Evaluation\label{app:sec:qualitative}}
\begin{figure*}
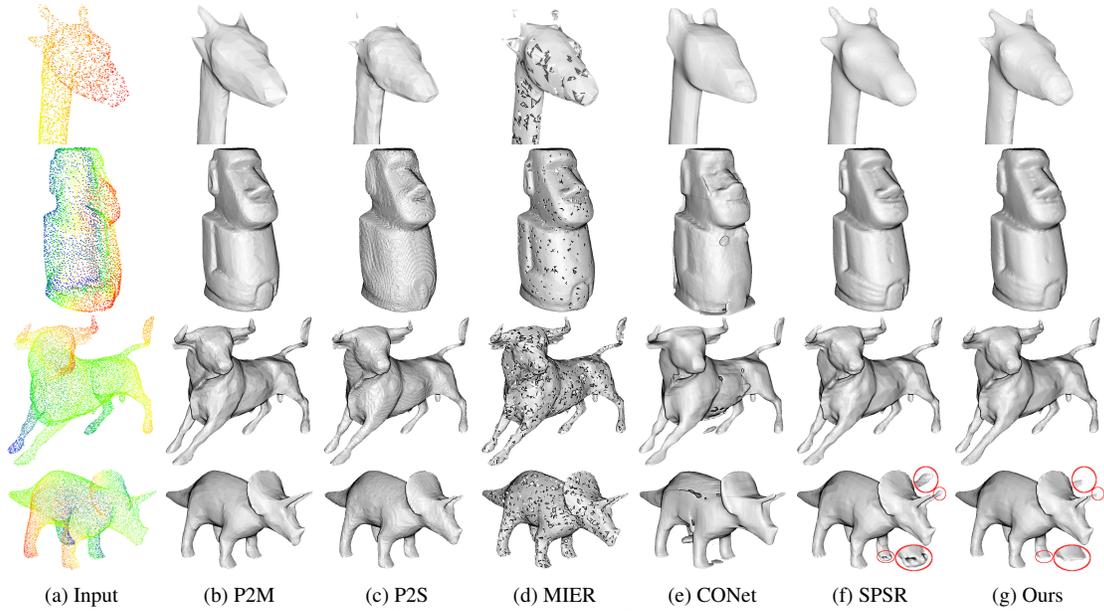

    \def\subimgwidth{0.12\linewidth}
    \edef\bullopts{trim=310 30 340 75,clip,width = \subimgwidth}
    \edef\triopts{trim=290 60 260 120,clip,width = \subimgwidth}
    \edef\grfopts{trim=370 0 510 15,clip,width = \subimgwidth}
    \edef\grfheadopts{trim=290 80 320 20,clip,width = \subimgwidth}
    \edef\tikiopts{trim=250 0 385 20,clip,width = \subimgwidth}
    \edef\DSLRViewOneopts{trim=200 100 300 50,clip,width = \subimgwidth}
    
    \centering
    \setlength{\tabcolsep}{0pt}
    \def\arraystretch{0}
    \begin{tabular}{ccccccc} 
    \subfloat[Input]{%
    \begin{tabular}{c}
         \expandafter\includegraphics\expandafter[\grfheadopts]{object_giraffe/pointcloud_ramp_view_2.jpg} \\
         \expandafter\includegraphics\expandafter[\tikiopts]{object_tiki/pointcloud_ramp_view_1.jpg} \\
         \expandafter\includegraphics\expandafter[\bullopts]{object_bull/pointcloud_ramp_view_1.jpg} \\
         \expandafter\includegraphics\expandafter[\triopts]{object_triceratops_raw/pointcloud_ramp_view_1.jpg} \\
    \end{tabular}} &
    \subfloat[P2M]{%
    \begin{tabular}{c}
         \expandafter\includegraphics\expandafter[\grfheadopts]{object_giraffe/point2mesh_view_2.jpg} \\
         \expandafter\includegraphics\expandafter[\tikiopts]{object_tiki/point2mesh_view_1.jpg} \\
         \expandafter\includegraphics\expandafter[\bullopts]{object_bull/point2mesh_view_1.jpg} \\ 
         \expandafter\includegraphics\expandafter[\triopts]{object_triceratops_raw/point2mesh_view_1.jpg} \\
    \end{tabular}} &
    \subfloat[P2S]{%
    \begin{tabular}{c}
         \expandafter\includegraphics\expandafter[\grfheadopts]{object_giraffe/p2s_view_2.jpg} \\
         \expandafter\includegraphics\expandafter[\tikiopts]{object_tiki/p2s_view_1.jpg} \\
         \expandafter\includegraphics\expandafter[\bullopts]{object_bull/p2s_view_1.jpg} \\ 
         \expandafter\includegraphics\expandafter[\triopts]{object_triceratops_raw/p2s_view_1.jpg} \\
    \end{tabular}} &
    \subfloat[MIER]{%
    \begin{tabular}{c}
         \expandafter\includegraphics\expandafter[\grfheadopts]{object_giraffe/mier_view_2.jpg} \\
         \expandafter\includegraphics\expandafter[\tikiopts]{object_tiki/mier_view_1.jpg} \\
         \expandafter\includegraphics\expandafter[\bullopts]{object_bull/mier_view_1.jpg} \\ 
         \expandafter\includegraphics\expandafter[\triopts]{object_triceratops_raw/mier_view_1.jpg} \\
    \end{tabular}} &
    \subfloat[CONet]{%
    \begin{tabular}{c}
        \expandafter\includegraphics\expandafter[\grfheadopts]{object_giraffe/convonet_view_2.jpg} \\
        \expandafter\includegraphics\expandafter[\tikiopts]{object_tiki/convonet_view_1.jpg} \\
        \expandafter\includegraphics\expandafter[\bullopts]{object_bull/convonet_view_1.jpg} \\
        \expandafter\includegraphics\expandafter[\triopts]{object_triceratops_raw/convonet_view_1.jpg} \\ 
    \end{tabular}} &
    \subfloat[SPSR]{%
    \begin{tabular}{c}
         \expandafter\includegraphics\expandafter[\grfheadopts]{object_giraffe/poisson_view_2.jpg} \\
         \expandafter\includegraphics\expandafter[\tikiopts]{object_tiki/poisson_view_1.jpg} \\
         \expandafter\includegraphics\expandafter[\bullopts]{object_bull/poisson_view_1.jpg} \\
         \expandafter\includegraphics\expandafter[\triopts]{object_triceratops_raw/poisson_view_1_mag.jpg} \\
    \end{tabular}} &
    \subfloat[Ours]{%
    \begin{tabular}{c}
        \expandafter\includegraphics\expandafter[\grfheadopts]{object_giraffe/ours_view_2.jpg} \\
        \expandafter\includegraphics\expandafter[\tikiopts]{object_tiki/ours_view_1.jpg} \\
        \expandafter\includegraphics\expandafter[\bullopts]{object_bull/ours_view_1.jpg} \\
        \expandafter\includegraphics\expandafter[\triopts]{object_triceratops_raw/ours_view_1_mag.jpg} \\
    \end{tabular}}
    \end{tabular}
\caption{Qualitative comparison on additional small object level datasets. From top to bottom, the data is \textit{Giraffe}, \textit{Tiki}, \textit{Bull}, and \textit{Triceratops}.}
\label{fig:qual:addsmallobjs}
\end{figure*}

\begin{figure*}
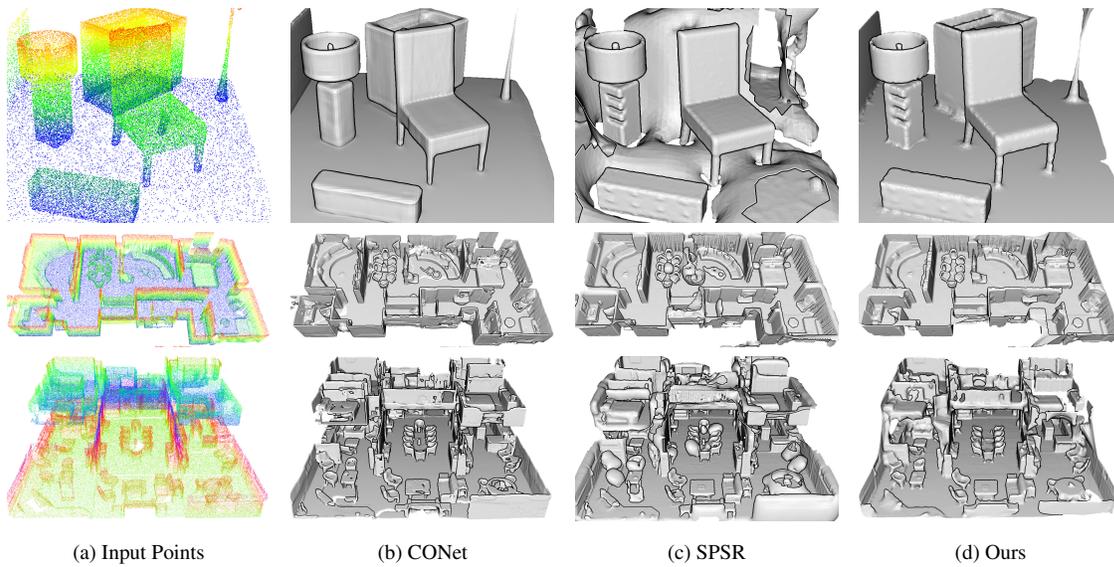

    \def\subimgwidth{0.2\linewidth}
    \edef\singleflooropts{trim=20 48 10 63,clip,width = \subimgwidth}
    \edef\multiflooropts{trim=100 10 200 50,clip,width = \subimgwidth}
    \edef\roomOneViewTwoopts{trim=300 30 230 50,clip,width = \subimgwidth}

    \centering
    \setlength{\tabcolsep}{2pt}
    \def\arraystretch{1}
    \begin{tabular}{cccc}
    \subfloat[Input Points]{%
    \begin{tabular}{c}
        \expandafter\includegraphics\expandafter[\roomOneViewTwoopts]{indoor_room_1_raw/pointcloud_ramp_view_2.jpg}\\
        \expandafter\includegraphics\expandafter[\singleflooropts]{indoor_matterport_singlefloor/pointcloud_ramp_view_1.jpg}\\
         \expandafter\includegraphics\expandafter[\multiflooropts]{indoor_matterport_multifloor/pointcloud_ramp_view_1.jpg}
    \end{tabular}} &
    \subfloat[CONet]{%
    \begin{tabular}{c}
        \expandafter\includegraphics\expandafter[\roomOneViewTwoopts]{indoor_room_1_raw/convonet_view_2.jpg} \\
        \expandafter\includegraphics\expandafter[\singleflooropts]{indoor_matterport_singlefloor/convonet_view_1.jpg} \\
         \expandafter\includegraphics\expandafter[\multiflooropts]{indoor_matterport_multifloor/convonet_view_1.jpg}
    \end{tabular}} &
    \subfloat[SPSR]{%
    \begin{tabular}{c}
        \expandafter\includegraphics\expandafter[\roomOneViewTwoopts]{indoor_room_1_raw/poisson_trim_view_2.jpg} \\
        \expandafter\includegraphics\expandafter[\singleflooropts]{indoor_matterport_singlefloor/poisson_trim_view_1.jpg} \\
         \expandafter\includegraphics\expandafter[\multiflooropts]{indoor_matterport_multifloor/poisson_trim_view_1.jpg}
    \end{tabular}} &
    \subfloat[Ours]{%
    \begin{tabular}{c}
        \expandafter\includegraphics\expandafter[\roomOneViewTwoopts]{indoor_room_1_raw/ours_angle_view_2.jpg} \\
        \expandafter\includegraphics\expandafter[\singleflooropts]{indoor_matterport_singlefloor/ours_view_1.jpg} \\
         \expandafter\includegraphics\expandafter[\multiflooropts]{indoor_matterport_multifloor/ours_view_1.jpg}
    \end{tabular}}
    \end{tabular}
    
    \caption{Qualitative comparison on additional indoor datasets. From top to bottom, the data is \textit{Room1}, \textit{MPT:SingleFloor}, and \textit{MPT:MultiFloor}.}
    \label{fig:qual:addindoor}
\end{figure*}

\begin{figure*}
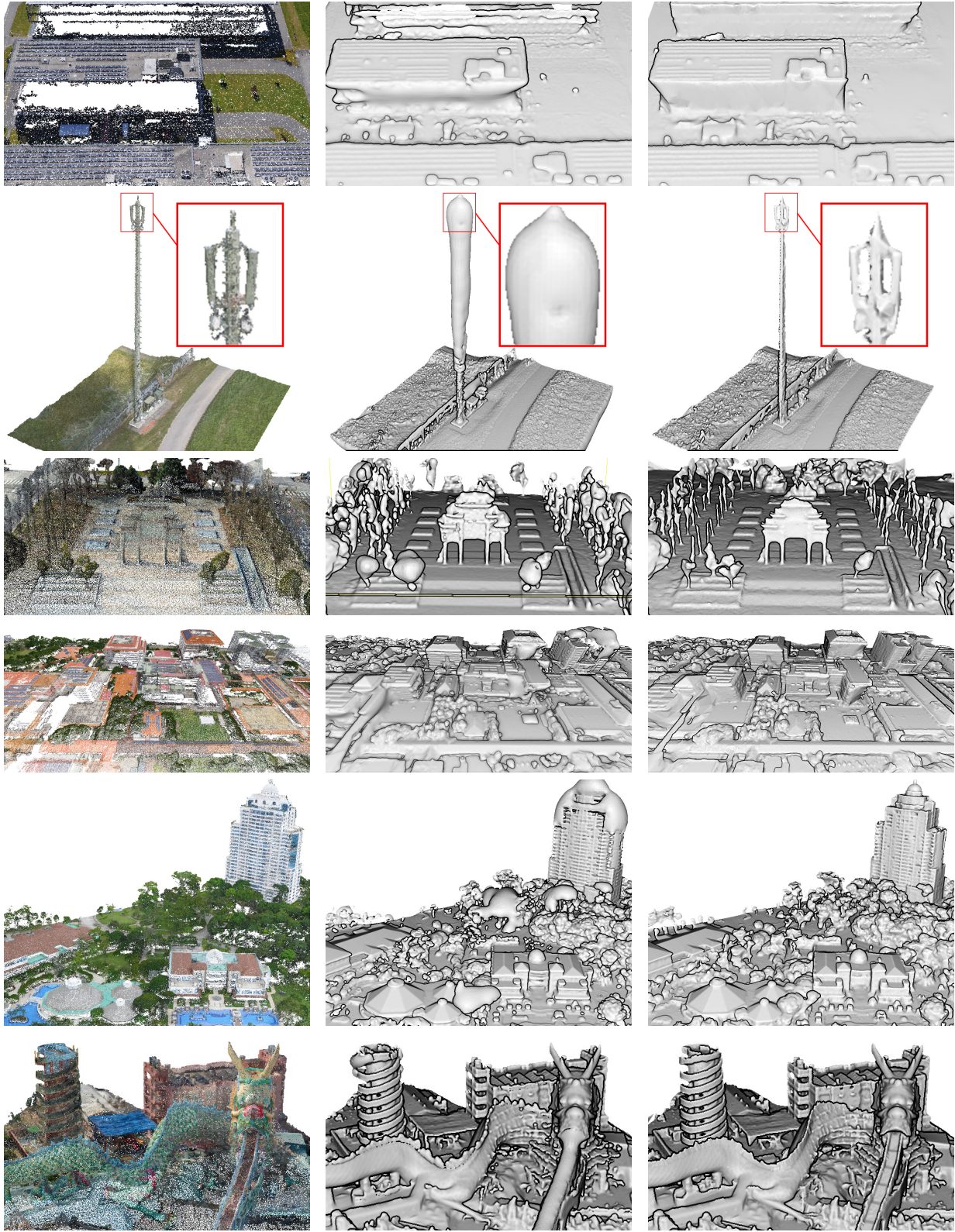

    \def\subimgwidth{0.3\linewidth}
    \edef\yparcopts{trim=235 100 300 125,clip,width = \subimgwidth}
    \edef\gsmopts{trim=200 0 200 0,clip,width = \subimgwidth}
    \edef\archwayopts{trim=100 80 150 220,clip,width = \subimgwidth}
    \edef\hotelopts{trim=100 20 300 10,clip,width = \subimgwidth}
    \edef\kittiopts{trim=100 20 300 10,clip,width = \subimgwidth}
    \edef\kittitwoopts{trim=5 20 5 160,clip,width = \subimgwidth}
    \edef\uthamopts{trim=50 10 50 100,clip,width = \subimgwidth}
    \edef\dragonparkopts{trim=250 200 450 110,clip,width = \subimgwidth}
    
    \centering
    \setlength{\tabcolsep}{2pt}
    \def\arraystretch{1}
    \begin{tabular}{ccc}
    \subfloat[Input Points]{%
    \begin{tabular}{c}
        \expandafter\includegraphics\expandafter[\yparcopts]{largescale_yparc/pointcloud.jpg} \\
        \expandafter\includegraphics\expandafter[\gsmopts]{largescale_gsm_tower/pointcloud_mag.jpg} \\
        \expandafter\includegraphics\expandafter[\archwayopts]{largescale_archway/pointcloud.jpg} \\
        \expandafter\includegraphics\expandafter[\uthamopts]{largescale_utham/pointcloud.jpg}\\
        \expandafter\includegraphics\expandafter[\hotelopts]{largescale_hotel/pointcloud_view_1.jpg} \\
        \expandafter\includegraphics\expandafter[\dragonparkopts]{dragonpark_5M/point_view1.jpg} \\
    \end{tabular}} &
    \subfloat[SPSR]{%
    \begin{tabular}{c}
        \expandafter\includegraphics\expandafter[\yparcopts]{largescale_yparc/poisson_trim.jpg} \\
        \expandafter\includegraphics\expandafter[\gsmopts]{largescale_gsm_tower/poisson_mag.jpg} \\
        \expandafter\includegraphics\expandafter[\archwayopts]{largescale_archway/poisson_trim.jpg} \\
        \expandafter\includegraphics\expandafter[\uthamopts]{largescale_utham/poisson_trim.jpg}\\
        \expandafter\includegraphics\expandafter[\hotelopts]{largescale_hotel/poisson_trim_view_1.jpg} \\
        \expandafter\includegraphics\expandafter[\dragonparkopts]{dragonpark_5M/poisson_trim_view1.jpg} \\
    \end{tabular}} &
    \subfloat[Ours]{%
    \begin{tabular}{c}
        \expandafter\includegraphics\expandafter[\yparcopts]{largescale_yparc/ours.jpg} \\
        \expandafter\includegraphics\expandafter[\gsmopts]{largescale_gsm_tower/ours_mag.jpg} \\
        \expandafter\includegraphics\expandafter[\archwayopts]{largescale_archway/ours.jpg} \\
        \expandafter\includegraphics\expandafter[\uthamopts]{largescale_utham/ours.jpg}\\
        \expandafter\includegraphics\expandafter[\hotelopts]{largescale_hotel/ours_view_1.jpg} \\
        \expandafter\includegraphics\expandafter[\dragonparkopts]{dragonpark_5M/ours_angle_view1.jpg} \\
    \end{tabular}}
    \end{tabular}
    
    \caption{Qualitative comparison on additional large scale datasets. From top to bottom, the data is \textit{YParc}, \textit{GSM Tower}, \textit{Archway}, \textit{UThammasat}, \textit{Hotel}, and \textit{Dragon Park}.}
    \label{fig:qual:addlargescale}
\end{figure*}

We present extra results qualitatively in three scales, for each, small objects (Figure~\ref{fig:qual:addsmallobjs}), indoor scenes (Figure~\ref{fig:qual:addindoor}), and large-scale outdoor datasets (Figure~\ref{fig:qual:addlargescale}).

{\small
\bibliographystyle{ieee_fullname}
\bibliography{egbib}
}